\documentclass{article}

\usepackage{microtype}
\usepackage{graphicx}
\usepackage{subcaption}
\usepackage{booktabs} 

\usepackage{hyperref}
\usepackage{adjustbox}

\usepackage[accepted]{icml2026}

\usepackage{amsmath}
\usepackage{amssymb}
\usepackage{mathtools}
\usepackage{amsthm}
\newcommand{\eg}{\emph{e.g.}}

\usepackage{pifont}
\usepackage{multirow}
\usepackage{multicol}

\usepackage[capitalize,noabbrev]{cleveref}

\theoremstyle{plain}

\theoremstyle{definition}

\theoremstyle{remark}

\usepackage[textsize=tiny]{todonotes}

\icmltitlerunning{Degradation-Aware Metric Prompting for Hyperspectral Image Restoration}

\begin{document}

\twocolumn[
  \icmltitle{Degradation-Aware Metric Prompting for Hyperspectral Image Restoration}

  \icmlsetsymbol{equal}{*}
  \icmlsetsymbol{corresp}{$\dagger$}

  \begin{icmlauthorlist}
    \icmlauthor{Binfeng Wang}{bit,zgc}
    \icmlauthor{Di Wang}{whu,zgc}
    \icmlauthor{Haonan Guo}{whu,zgc,corresp}
    \icmlauthor{Ying Fu}{bit,corresp}
    \icmlauthor{Jing Zhang}{whu,zgc,corresp}
  \end{icmlauthorlist}

  \icmlaffiliation{bit}{School of Computer Science and Technology, Beijing Institute of Technology, Beijing, China}
  \icmlaffiliation{whu}{School of Computer Science, Wuhan University, Wuhan, Hubei, China}
  \icmlaffiliation{zgc}{Zhongguancun Academy, Beijing, China }

  \icmlcorrespondingauthor{Haonan Guo}{haonan.guo@whu.edu.cn}
  \icmlcorrespondingauthor{Ying Fu}{fuying@bit.edu.cn}
  \icmlcorrespondingauthor{Jing Zhang}{jingzhang.cv@gmail.com}

  \icmlkeywords{Unified hyperspectral image restoration, Degradation-aware metric}

  \vskip 0.3in
]

\printAffiliationsAndNotice{$\dagger$ Corresponding author}

\begin{abstract}

Unified hyperspectral image (HSI) restoration aims to recover diverse degradations within a single model. However, current methods often rely on impractical explicit priors or opaque black-box representations that overfit to training distributions, hampering generalization to unseen scenarios. To bridge this gap, we propose Degradation-Aware Metric Prompting (DAMP), a novel framework that characterizes multi-dimensional degradations through interpretable spatial-spectral metrics. These metrics serve as Degradation Prompts (DP), enabling the model to capture shared characteristics across tasks and adapt to unknown corruptions. Central to our framework is the Degradation-Adaptive Mixture-of-Experts (DAMoE), where Spatial-Spectral Adaptive Modules (SSAMs) serve as experts that utilize learnable fusion coefficients to specialize in distinct degradation degrees. By using DP as a gating router, DAMoE dynamically activates specialized experts tailored to the specific degradation profile. Extensive experiments on natural and remote sensing HSI datasets demonstrate that DAMP achieves state-of-the-art performance and exhibits exceptional zero-shot generalization on unseen restoration tasks. Code is publicly available at \href{DAMP}{https://github.com/MiliLab/DAMP}.
\end{abstract}

\section{Introduction}

Hyperspectral images (HSIs) capture spectral information across hundreds of contiguous bands. However, practical acquisition is often degraded by physical and hardware limitations, such as low signal-to-noise ratio, motion blur, or calibration errors~\cite{rasti2021degsurvey1,zhang2013degsurvey2}. These degradations weaken the discriminative spectral features essential for material identification, significantly hindering downstream tasks like classification~\cite{audebert2019HSICls} and change detection~\cite{liu2019HSIChange} in applications including agriculture~\cite{agri} and environmental monitoring~\cite{Env}.

\begin{figure}[t]
    \begin{center}
    \centerline{\includegraphics[width=\linewidth]{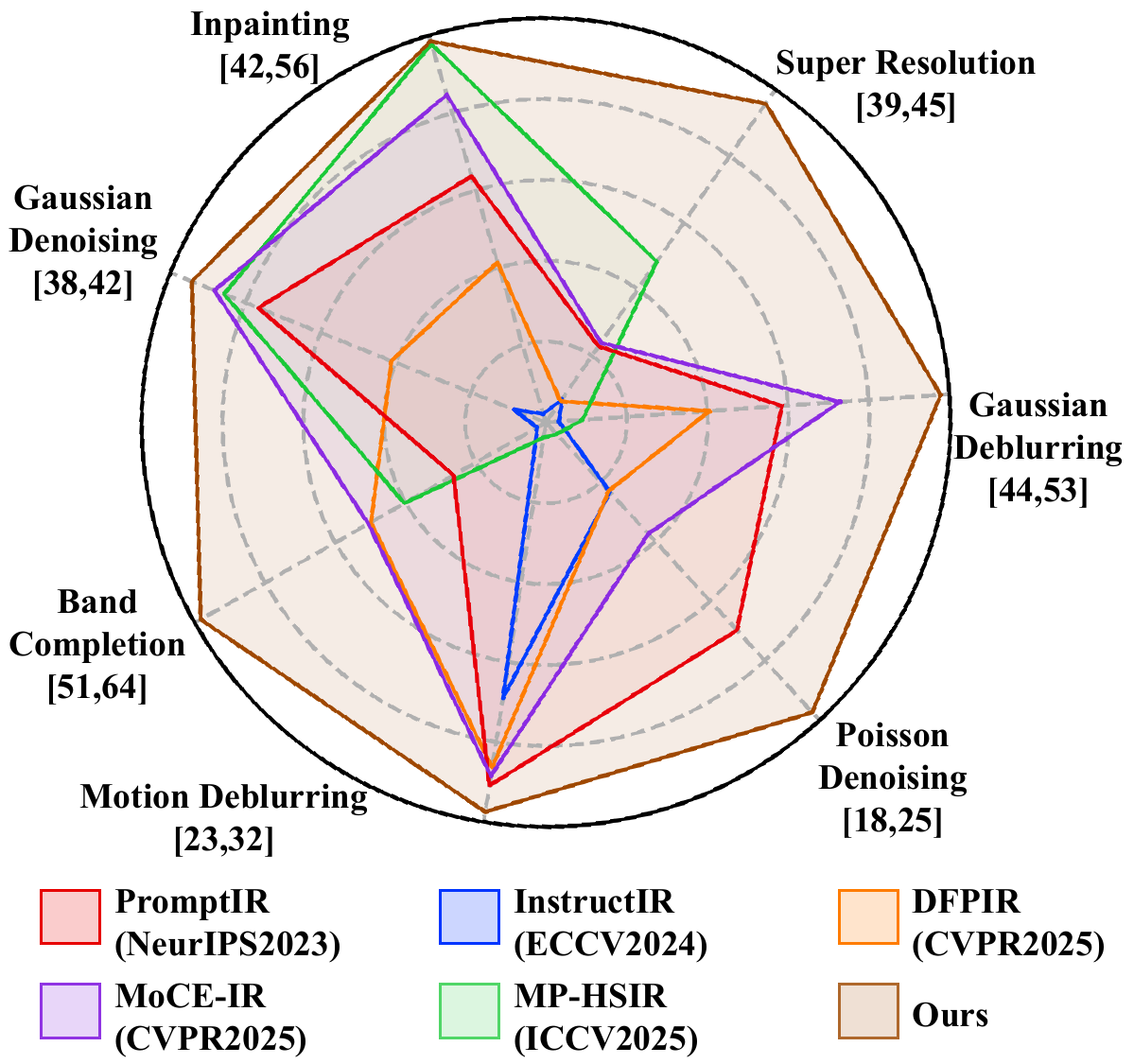}}
    \caption{PSNR comparison with the SOTA all-in-one methods: Inpainting, Super Resolution, Gaussian Deblurring, and Gaussian Denoising results are evaluated on the ARAD dataset after unified training, while Poisson Denoising and Motion Deblurring are reported as zero-shot results on the CAVE dataset. $[\cdot]$ denotes the range of PSNR values across different methods.}
    \label{fig1}
    \end{center}
\end{figure}

To address these issues, specialized methods have been developed for individual degradation types, such as denoising~\cite{li2023SST,zhang2026real} and super-resolution~\cite{li2017hsiSR, li2023hyperspectral, zhang2025unaligned}. However, real-world HSI degradations are often complex and not easily characterized a prior, involving mixtures of noise, blur, and band missing~\cite{lee2024PromptHSI}. Inspired by prompt-based frameworks in natural image restoration~\cite{potlapalli2023PromptIR, conde2024InstructIR, wang2025moerl}, several recent HSI methods have adopted similar paradigms to handle diverse degradations within a single model~\cite{lee2024PromptHSI, wu2025MP-HSIR}.

Despite their progress, current prompt-based approaches remain constrained by their reliance on either explicit degradation priors (\eg, predefined types or text prompts)~\cite{lee2024PromptHSI, wu2025MP-HSIR} or implicit black-box representations (\eg, Convolutional Neural Networks (CNN) embeddings)~\cite{potlapalli2023PromptIR, wang2025moerl}. This reliance introduces two critical limitations: (i) Inflexibility in Complex Scenarios: Explicit priors assume a closed set of degradations, failing to characterize the complex, mixed, and continuous degradation patterns found in real-world HSIs. (ii) Limited Generalization and Spectral Fidelity: Black-box models tend to overfit the degradation distribution of training data, leading to poor generalization on unseen corruptions. Moreover, they typically lack explicit mechanisms to model spectral correlations, resulting in suboptimal spectral recovery.

In this work, we propose Degradation-Aware Metric Prompting (DAMP), a unified HSI restoration framework that eliminates the need for explicit degradation priors or implicit black-box representations. Instead of relying on predefined degradation types, textual prompts, or opaque latent vectors, we quantify degradations using interpretable spatial-spectral metrics. Our analysis indicates that (i) a small set of key metrics can effectively distinguish typical degradation types, and (ii) distinct degradation types can exhibit overlapping distributions across specific metrics (\eg, slight blur and low noise), reflecting underlying commonalities despite their different physical origins. Based on these observations, we introduce Degradation Prompts (DP), which continuously encode degradation severity across dimensions to naturally reflect task commonality.
To translate these degradation insights into adaptive restoration, we propose the Degradation-Adaptive Mixture-of-Experts (DAMoE) architecture. Within this framework, Spatial-Spectral Adaptive Modules (SSAMs) serve as specialized experts, utilizing learnable fusion coefficients to dynamically modulate spatial and spectral feature extraction for distinct degradation degrees. By employing the DP as a degradation-aware gating router, DAMoE intelligently activates and aggregates the optimal experts tailored to the specific input. This design enables adaptive and unified restoration under diverse, mixed, or unseen degradations. As shown in Fig.~\ref{fig1}, extensive experiments demonstrate that DAMP achieves state-of-the-art (SOTA) performance across multiple HSI restoration tasks and generalizes effectively to unseen degradation scenarios.
Our main contributions are summarized as follows:

\begin{itemize}
    \item We propose DAMP, a unified HSI restoration framework that eliminates the need for explicit degradation priors while offering an interpretable characterization of degradations. By leveraging degradation-aware metric representations, DAMP adaptively handles diverse, mixed, and unseen degradations within a single model.

    \item We introduce DP that continuously quantify multi-dimensional degradations using spatial-spectral metrics. DP effectively captures degradation severity and shared characteristics across restoration tasks.

    \item We design DAMoE, where DP acts as a degradation-aware gating router and SSAM modules serve as specialized experts, enabling adaptive restoration under complex and unknown degradations.
    
\end{itemize}

\section{Related Work}

\subsection{Unified Natural Image Restoration} 
Unified image restoration (UIR) aims to handle diverse degradations within a single model, enabling flexible adaptation without task-specific architectures. To achieve this, recent approaches have explored various conditioning mechanisms to guide network behavior based on degradation types. Some employ prompts or natural language instructions to steer restoration processes~\cite{potlapalli2023PromptIR, conde2024InstructIR}, drawing inspiration from advances in vision-language modeling. Others leverage diffusion models for strong generalization through shared priors or iterative refinement~\cite{zheng2024DiffUIR, zhou2025UniRes, li2025LD-RPS}, while complementary strategies enhance robustness via synthetic degradation generation~\cite{rajagopalan2025GenDeg} or large-scale training on real-world data~\cite{li2024FoundIR}. 
Architectural innovations further improve performance and efficiency, including multi-stage refinement~\cite{zamir2021MPRNet}, degradation-aware encoding~\cite{li2022AirNet}, feature perturbation for task disentanglement~\cite{tian2025DFPIR}, frequency-aware progressive recovery for high-resolution inputs~\cite{liu2025UHD}, visual instruction-based guidance~\cite{luo2025VIDIR}, and MoE designs with dynamic routing~\cite{zamfir2025MoCEIR, wang2025moerl}.
While these methods demonstrate promising generalization in the natural image domain, they are primarily designed for RGB data and do not account for the complex spectral-spatial degradation of HSIs. 
In contrast, our DP captures degradation patterns across spatial-spectral dimensions, and our DAMoE enables adaptive restoration tailored to varying degrees of spatial and spectral degradation.

\subsection{Hyperspectral Image Restoration} 
HSI restoration has evolved from traditional model-based optimization to deep learning-driven solutions. Early methods relied on handcrafted priors such as sparsity~\cite{rasti2014sparse, rasti2017sparse, liu2018sparse}, non-local self-similarity~\cite{xu2017nonlocal, qian2012nonlocal}, and total variation~\cite{wang2017totalvaria, he2015totalvaria} to regularize the ill-posed inverse problem. With the rise of data-driven approaches, convolutional neural networks~\cite{chang2018hsiCNN, zeng2020hsiCNN, zhang2024deep}, transformers~\cite{long2023hsitrans, yu2023hsitrans, wang2026visual}, and more recently diffusion models~\cite{li2024hsidiff, pang2024hsidiff, chen2026any2any} have demonstrated superior performance by learning implicit image priors from large-scale datasets.
Despite these advances, most existing HSI restoration methods are task-specific, such as denoising~\cite{li2023SST, li2025noise} or super-resolution~\cite{li2017hsiSR, wang2023hsiSR, zhang2026enhancing}, and lack flexibility in handling diverse or mixed degradations within a unified framework. 
Inspired by UIR in the natural image domain, recent efforts adopt conditional mechanisms.
For example, PromptHSI~\cite{lee2024PromptHSI} and MP-HSIR~\cite{wu2025MP-HSIR} use multi-modal prompts or explicit degradation labels to enable task-adaptive, instruction-guided restoration.
Yet, such approaches rely on external degradation labels that often impractical to obtain, limiting their real-world applicability. 
Our method derives interpretable spatial-spectral metrics directly from the input to characterize degradations, enabling autonomous and unified restoration independent of external degradation labels.

\begin{figure*}[t]
    \begin{center}
    \centerline{\includegraphics[width=\linewidth]{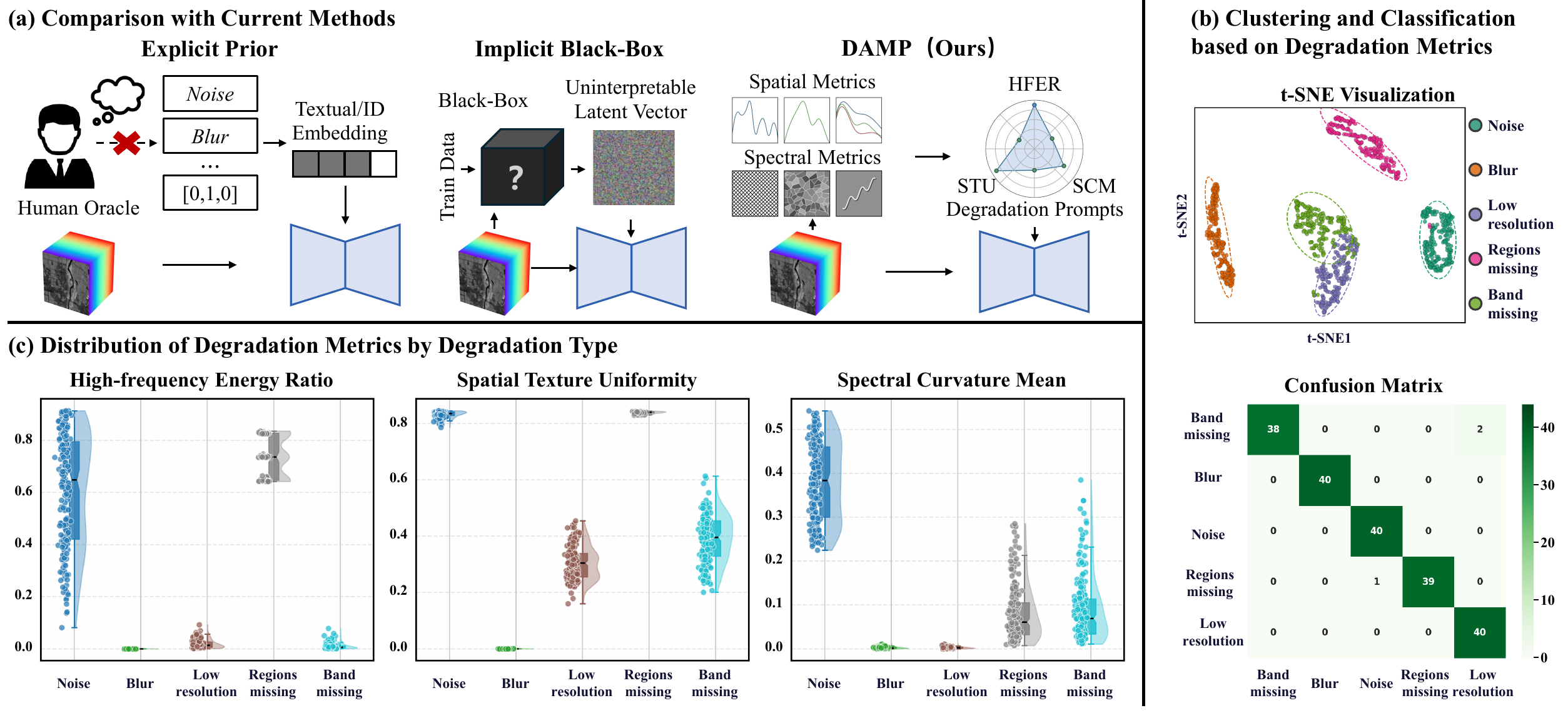}}
    \caption{(a) Comparison between explicit degradation priors-based methods, implicit black-box representation-based method and degradation-aware metric prompting approaches. (b) Confusion matrix for classifying five degradation types based on HFER, STU and SCM. (c) Distribution of different degradation types across the HFER, STU and SCM. }
    \label{fig2}
    \end{center}
\end{figure*}

\section{Method}

In this section, we present DAMP, a Degradation-Aware Metric Prompting framework for unified HSI restoration that operates without explicit degradation priors or implicit black-box representations. Section~\ref{sec:task} formally defines the unified HSI restoration task. Section~\ref{sec:dp} then investigates the relationship between measurable degradation metrics and restoration requirements, leading to the proposal of DP. Finally, Section~\ref{sec:damp} introduces the overall pipeline of DAMP framework and its key component DAMoE.

\begin{figure*}[ht]
    \begin{center}
    \centerline{\includegraphics[width=0.9\linewidth]{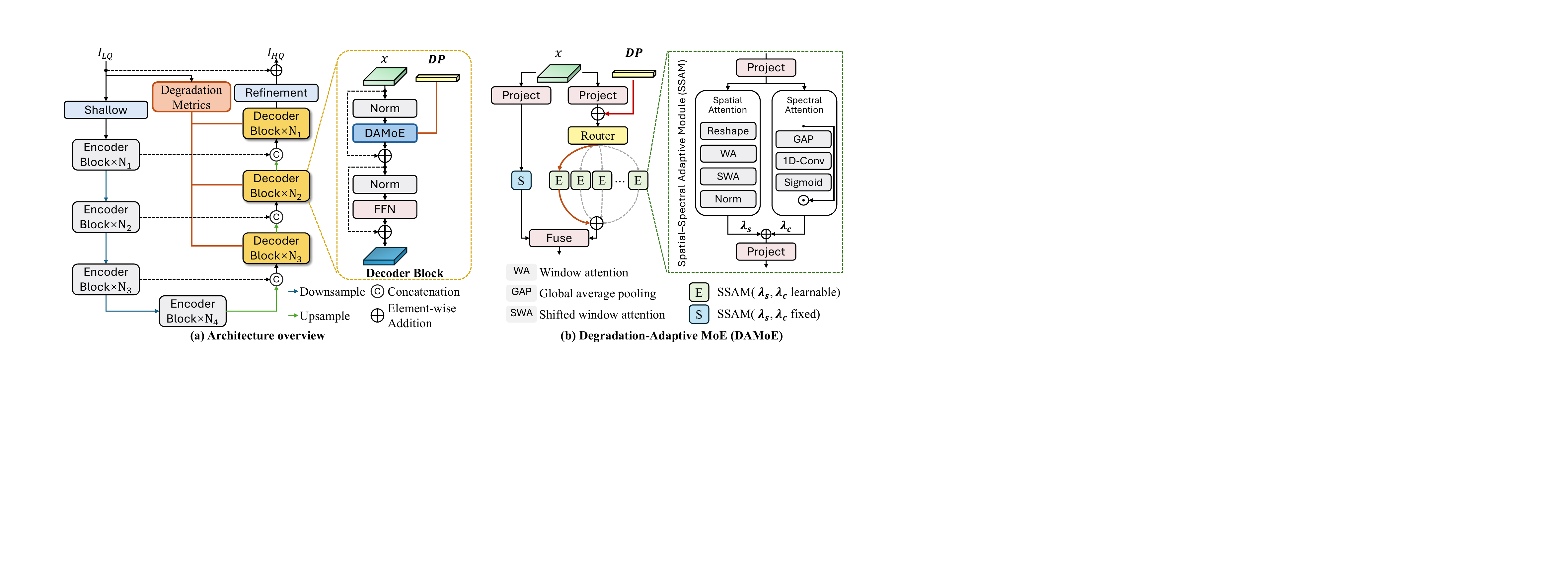}}
    \caption{(a) The architecture of the proposed DAMP framework. (b) The Degradation-Adaptive MoE.}
    \label{fig:framework}
    \end{center}
\end{figure*}

\begin{table*}[t]

\caption{Definitions and physical interpretations of spatial and spectral degradation metrics.}
\label{tab:metrics}
\begin{adjustbox}{max width=0.9\textwidth, center}
\begin{tabular}{l l l}
\toprule
\textbf{Metric} & \textbf{Mathematical Formulation} & \textbf{Physical Interpretation} \\
\midrule
HFER & $ \frac{1}{C} \sum_{c=1}^{C} \frac{\sum_{(u,v) \in \Omega_H} |\mathcal{F}[x_c(u,v)]|^2}{\sum_{(u,v)} |\mathcal{F}[x_c(u,v)]|^2}$ & Proportion of high-frequency energy indicating spatial detail preservation \\
STU & $\frac{1}{C} \sum_{c=1}^{C} \frac{\exp\left(\frac{1}{HW} \sum_{h,w} \ln|\mathcal{F}[x_c(h,w)]|\right)}{\frac{1}{HW} \sum_{h,w} |\mathcal{F}[x_c(h,w)]|}$ & Spectral smoothness measured by geometric-to-arithmetic mean ratio \\
SCM & $ \frac{1}{C-2} \sum_{i=1}^{C-2} |\nabla^2 s_i|$ & Average absolute curvature of spectral curves indicating spectral continuity \\
\bottomrule
\end{tabular}
\end{adjustbox}
\end{table*}

\subsection{Problem Formulation}
\label{sec:task}

Unified HSI restoration aims to recover the clean image $\mathcal{X}$ from an observed degraded image $\mathcal{Y}$ that may suffer from arbitrary types of degradation. Formally, this can be expressed as the following inverse problem:
\begin{equation}
    \mathcal{Y} = \mathcal{D}(\mathcal{X}) + \mathcal{N},
\end{equation}
where $\mathcal{D}(\cdot)$ represents an arbitrary degradation operator, which may include blur, missing stripes, compression artifacts, or any combination thereof, and $\mathcal{N}$ denotes additive noise (\eg, Gaussian noise).

The goal of unified restoration is to learn a restoration mapping $\mathcal{R}_\theta(\cdot)$ such that
\begin{equation}
    \hat{\mathcal{X}} = \mathcal{R}_\theta(\mathcal{Y}),
\end{equation}
where $\hat{\mathcal{X}}$ denotes the restored clean HSI, and the model is expected to generalize across diverse and mixed degradation types within a single unified framework.

\subsection{Degradation Prompts}
\label{sec:dp}

Characterizing HSI degradation in a unified manner remains an open challenge. As illustrated in Fig.~\ref{fig2} (a), existing methods typically face significant trade-offs: they either rely on explicit degradation priors~\cite{wu2025MP-HSIR, lee2024PromptHSI}, which are often difficult to obtain accurately in real-world scenarios, or employ implicit black-box representations~\cite{wang2025moerl,potlapalli2023PromptIR} that are uninterpretable and prone to overfitting training distributions, leading to poor generalization on unseen degradations. To address these limitations, we introduce DP, a compact multi-dimensional representation grounded in interpretable metrics. Unlike previous approaches, DP operates without explicit priors and effectively generalizes to unseen degradation types.

\noindent \textbf{Feasibility Analysis with Representative Metrics.}
To validate whether simple interpretable metrics can effectively characterize complex degradations, we initially investigate three representative metrics: High-Frequency Energy Ratio (HFER) and Spatial Texture Uniformity (STU) for spatial degradation, and Spectral Curvature Mean (SCM) for spectral continuity. The mathematical formulations and physical interpretations of these metrics are detailed in Table~\ref{tab:metrics}.

We analyze these metrics on a pilot dataset of 1,000 degraded HSIs. As shown in Fig.~\ref{fig2} (b), the t-SNE visualization reveals that these three metrics alone can form distinct clusters for different degradation types, while the confusion matrix obtained from a Random Forest classifier further verifies their strong discriminative capability. Furthermore, the distribution analysis in Fig.~\ref{fig2} (c) uncovers critical insights: distinct degradation types (\eg, regions missing and band missing) can exhibit overlapping distributions in specific metrics (\eg, SCM). These findings demonstrate that interpretable metrics can capture both the identity and the underlying commonality of degradations. Crucially, this interpretability fundamentally enhances generalization. Unlike black-box methods that implicitly encode inputs into latent prompts bound to the training distribution, often projecting unseen degradations onto the learned manifold of known types, DP explicitly decomposes degradations into intrinsic spatial-spectral indicators. By objectively quantifying the physical degree of degradation independent of data bias, DP offers a universal representation that remains faithful even in unseen scenarios.

\noindent \textbf{Optimal Metric Selection for DAMP.}
Building on these observations, we formalize a selection pipeline to construct the final DP vector with maximal discriminative power. Starting from a comprehensive pool of 25 candidate metrics covering entropy, gradient, and frequency statistics, we apply a systematic three-stage filtering process: (1) Interpretability, where we filter out abstract statistical features lacking clear physical correspondence to degradation mechanisms; (2) Spatial-Spectral Coverage, ensuring the retained set jointly represents both spatial structure and spectral fidelity; and (3) Discriminability, where we select the final metrics based on their feature importance scores derived from the Random Forest degradation classification, ensuring the retention of the most informative indicators for distinguishing degradation types.

This rigorous process yields six final metrics that form the input to our DAMP framework: (1) High-Frequency Energy Ratio, (2) Spatial Texture Uniformity, (3) Spectral Curvature Mean, (4) Spectral Curvature Standard Deviation, (5) Gradient Standard Deviation, and (6) Spatial Correlation Coefficient. This 6-dimensional DP vector provides a robust, lightweight, and transparent descriptor of the degradation state. The detailed selection pipeline and definitions of all six metrics are provided in Supplementary Section A.

\subsection{DAMP Framework}
\label{sec:damp}

\textbf{Overall Pipeline.} 
As illustrated in Fig.~\ref{fig:framework} (a), our pipeline adopts a hierarchical U-shaped architecture tailored for efficiency, guided by the proposed DP. 
Initially, degradation metrics are extracted from the degraded input and projected into a metric embedding space to form the DP vector. Concurrently, a $3\times3$ convolution extracts shallow features from the input, which are then processed through four hierarchical encoding and decoding stages.
While the encoder utilizes standard attention blocks to extract multi-scale features, the core innovation lies in the decoder, where we introduce the DAMoE. 
Crucially, this integration occurs across all hierarchical levels of the decoder. This allows the DP to act as a global condition that dynamically modulates the restoration trajectory, guiding the recovery from coarse semantic features to fine-grained spatial-spectral details. 
Finally, a residual block fuses the original input with the enhanced features from the decoder to produce the restored image.

\textbf{DAMoE.} 
To enable the model to adaptively select an appropriate restoration strategy based on the specific degradation pattern, we propose the DAMoE, as shown in Fig.~\ref{fig:framework} (b).
Unlike previous MoE approaches that rely on abstract visual features or implicit black-box prompts~\cite{wang2025moerl} for routing, DAMoE incorporates the explicit DP vector $\mathbf{e}$ as a guiding prior to steer the expert selection.

Formally, given an input feature map $\mathbf{x}$, it is processed by a set of experts. The gating score $\mathbf{g}$ for selecting these experts is computed via a degradation-aware routing function:
\begin{equation}
    \mathbf{g} = \mathcal{T}_{k} \left( \text{softmax} \left( \mathbf{W}_g \cdot \sigma \left( \mathbf{W}_{proj} [ \text{GAP}(\mathbf{x}), \mathbf{e} ] \right) + \mathbf{\epsilon} \right) \right),
\end{equation}
where $\text{GAP}(\cdot)$ denotes Global Average Pooling to align the spatial dimensions of the features with the DP vector, $[\cdot, \cdot]$ represents the concatenation operation, and $\mathbf{W}_{proj}$ and $\mathbf{W}_g$ are learnable projection matrices. $\sigma$ denotes the activation function. 
To promote load balancing and expert exploration during training, we add Gaussian noise $\mathbf{\epsilon} \sim \mathcal{N}(0, 1)$ to the logits before the softmax. $\mathcal{T}_{k}(\cdot)$ is the top-$k$ sparsification operator that retains only the $k$ largest entries.

This design ensures that the routing logic is explicitly grounded in the physical properties of the degradation. For instance, an input with a high HFER value (indicating severe noise) will explicitly bias the router towards experts with stronger spectral filtering capabilities, even if the visual features are ambiguous.
The final degradation-aware feature representation is computed as the weighted sum of the selected experts:
\begin{equation}
    \mathbf{f}_{deg} = \sum_{i \in \mathcal{K}} g_i \cdot \mathbf{f}_i,
\end{equation} 
where $g_i$ is the normalized gating score for the $i$-th expert, and $\mathcal{K}$ is the set of selected top-$k$ indices. Finally, the degradation-agnostic features extracted by shared experts are fused with $\mathbf{f}_{deg}$ via a channel-wise convolution to produce the output.

\textbf{SSAM.} 
HSIs frequently exhibit complex degradation patterns characterized by heterogeneous distortions across spatial and spectral dimensions (\eg, blur degrades spatial texture while preserving spectral curves, whereas noise affects both). Standard modules often enforce a static trade-off between these dimensions. 
To overcome this limitation, we propose the SSAM, designed as a specialized operator with expert-specific preferences.

Each SSAM expert employs parallel branches to extract features. The output of the $i$-th expert is generated via:
\begin{equation}
    \mathbf{F}_{expert}^{(i)} = \lambda_{s}^{(i)}\mathcal{E}_{s}(\mathbf{F}) + \lambda_{c}^{(i)}\mathcal{E}_{c}(\mathbf{F}), \quad \text{s.t. } \lambda_{s}^{(i)} + \lambda_{c}^{(i)} = 1,
\end{equation}
where $\mathcal{E}_{s}(\cdot)$ utilizes Window-based Multi-head Self-Attention (W-MSA) to capture spatial structural dependencies, and $\mathcal{E}_{c}(\cdot)$ employs 1D convolutions to model inter-spectral correlations.
We intentionally employ concise, streamlined architectures for these branches. This design ensures computational efficiency and shifts the modeling complexity from individual operators to the dynamic orchestration of experts, allowing the framework to tackle complex degradations through the adaptive composition of fundamental spatial-spectral bases.
A key design choice is that $\lambda_{s}^{(i)}$ and $\lambda_{c}^{(i)}$ are \textit{expert-specific learnable parameters}, rather than instance-specific predictions. 
This constraint forces distinct experts to ``specialize'' during training, creating ``Spatial Experts'' (high $\lambda_{s}$) that prioritize texture recovery and ``Spectral Experts'' (high $\lambda_{c}$) that focus on spectral fidelity. 
Consequently, the router can dynamically assemble the optimal restoration strategy by mixing these specialized experts within our DAMoE according to the degradation severity indicated by the DP.

\begin{table*}[ht]

\caption{Quantitative comparison on five HSI restoration tasks across natural and remote sensing datasets.}
\label{tab:seen_comparison}

\begin{adjustbox}{max width=\textwidth, center}
\begin{tabular}{c|c|c|c|c|c|c|c|c}
\hline
\multirow{3}{*}{\textbf{Type}} &
\multirow{3}{*}{\textbf{Method}} &
\multicolumn{3}{c|}{\textbf{ Gaussian Deblurring (Radius = 9, 15)}} &
\multirow{3}{*}{\textbf{Method}} &
\multicolumn{3}{c}{\textbf{Super-Resolution (Scale = 2, 4)}} \\
\cline{3-5} \cline{7-9}
& & PaviaU & ARAD & HyRank & & ICVL & ARAD & HyRank \\
\cline{3-5} \cline{7-9}
& & PSNR / SSIM / SAM & PSNR / SSIM / SAM & PSNR / SSIM / SAM & &PSNR / SSIM / SAM &PSNR / SSIM / SAM &PSNR / SSIM / SAM\\
\hline
\multirow{3}{*}{Task Specific}
& Stripformer & 30.14 / 0.861 / 5.103 & 48.41 / 0.995 / 0.883 & 42.31 / 0.942 / 6.673 & ESSA & 45.30 / 0.976 / 0.975 & 40.36 / 0.964 / 1.176 & 48.34 / 0.979 / 2.843\\
& DeepRFT & 31.91 / 0.901 / 5.221 & 46.28 / 0.993 / 0.923 & 44.22 / 0.969 / 3.224 & SRDNet & 44.55 / 0.974 / 1.027 & 39.59 / 0.959 / 1.812 & 48.11 / 0.976 / 3.091 \\
& Loformer & 32.42 / 0.909 / 4.257 & 49.91 / 0.997 / 0.762 & 45.33 / 0.972 / 3.053 & VolFormer & 44.71 / 0.975 / 0.998 & 41.83 / 0.970 / 1.083 & 48.91 / 0.983 / 2.245 \\
\hline
\multirow{6}{*}{All in One} 
& InstructIR & 30.02 / 0.857 / 5.264 &  44.04 / 0.987 / 1.321 & 45.56 / 0.977 / 2.835 & InstructIR & 44.09 / 0.973 / 1.445 & 39.82 / 0.964 / 1.613 & 48.63 / 0.981 / 2.449\\
& PromptIR & 32.21 / 0.908 / 5.141 & 49.18 / 0.996 / 0.822 & 46.12 / 0.982 / 2.157 & PromptIR & 45.44 / 0.978 / 0.956 & 40.57 / 0.966 / 1.168 & 48.82 / 0.981 / 2.768 \\
& DFPIR & 32.38 / 0.903 / 4.303 & 47.51 / 0.994 / 0.908 & 41.73 / 0.957 / 4.015 & DFPIR & 44.96 / 0.976 / 1.007 & 40.03 / 0.961 / 1.216 & 48.70 / 0.982 / 2.314 \\
& MoCE-IR & 31.68 / 0.896 / 4.598 & 50.52 / 0.996 / 0.673 & 42.47 / 0.947 / 6.521 & MoCE-IR & 46.04 / \textbf{0.981} / 0.896 & 40.62 / 0.967 / 1.110 & 44.17 / 0.924 / 6.651 \\
& MP-HSIR & 31.54 / 0.891 / 5.325 & 44.58 / 0.984 / 0.900 & 41.54 / 0.929 / 4.470 & MP-HSIR & 45.86 / 0.980 / 1.016 & 41.77 / 0.972 / 1.142 & 48.73 / 0.981 / 2.574 \\
& Ours & \textbf{33.84} / \textbf{0.929} / \textbf{4.246} & \textbf{52.84} / \textbf{0.998} / \textbf{0.508} & \textbf{46.39} / \textbf{0.984} / \textbf{2.112} & Ours & \textbf{46.33} / 0.980 / \textbf{0.882} & \textbf{44.01} / \textbf{0.981} / \textbf{0.866} & \textbf{49.77} / \textbf{0.987} / \textbf{2.165} \\

\hline

\multirow{3}{*}{\textbf{Type}} &
\multirow{3}{*}{\textbf{Method}} &
\multicolumn{3}{c|}{\textbf{ Inpainting (MaskRate = 0.7, 0.8, 0.9)}} &
\multirow{3}{*}{\textbf{Method}} &
\multicolumn{3}{c}{\textbf{ Gaussian Denoising (Sigma = 30, 50, 70)}} \\
\cline{3-5} \cline{7-9}
& & PaviaC & Xiong’an & Chikusei & & ICVL & ARAD & PaviaC \\
\cline{3-5} \cline{7-9}
& & PSNR / SSIM / SAM & PSNR / SSIM / SAM & PSNR / SSIM / SAM & &PSNR / SSIM / SAM &PSNR / SSIM / SAM &PSNR / SSIM / SAM\\
\hline
\multirow{3}{*}{Task Specific}
& SSDL & 27.97 / 0.722 / 16.114 & 29.43 / 0.534 / 15.561 & 35.43 / 0.823 / 14.224 & SERT & 41.03 / 0.961 / 4.441 & 40.27 / 0.959 / 3.891 & 19.11 / 0.301 / 35.124 \\
& Restormer & 29.35 / 0.784 / 13.764 & 33.11 / 0.675 / 12.934 & 33.51 / 0.792 / 16.345  & SST & 41.56 / 0.966 / 4.047 & 40.54 / 0.961 / 3.996 & 22.41 / 0.443 / 33.693 \\
& PFGIN & 28.64 / 0.734 / 14.549 & 32.34 / 0.629 / 13.346 & 31.42 / 0.734 / 18.821 & HCANet & 42.32 / 0.971 / 2.786 & 41.39 / \textbf{0.968} / 2.898 & 23.72 / 0.514 / 31.638 \\
\hline
\multirow{6}{*}{All in One} 
& InstructIR & 21.91 / 0.512 / 22.690 & 22.05 / 0.215 / 18.292 & 30.63 / 0.682 / 19.834 & InstructIR & 39.78 / 0.908 / 5.114 & 38.47 / 0.905 / 6.843 & 16.04 / 0.189 / 38.318 \\
& PromptIR & 27.56 / 0.716 / 16.660 & 31.36 / 0.579 / 13.600 & 37.52 / 0.896 / 12.642 & PromptIR & 42.35 / 0.970 / 2.659 & 40.85 / 0.963 / 3.548 & 23.22 / 0.501 / 32.128\\
& DFPIR & 22.41 / 0.555 / 21.606 & 22.70 / 0.252 / 18.268 & 31.39 / 0.723 / 18.778 & DFPIR & 41.11 / 0.957 / 4.370 & 39.61 / 0.948 / 5.745 & 17.74 / 0.279 / 39.633 \\
& MoCE-IR & 26.06 / 0.681 / 18.804 & 29.04 / 0.518 / 15.793 & 34.97 / 0.817 / 15.544 & MoCE-IR & 42.66 / 0.973 / 2.434 & 41.26 / 0.966 / 2.974 & 22.49 / 0.471 / 34.772 \\
& MP-HSIR & 29.29 / 0.775 / 14.106 & 33.42 / 0.697 / 11.129 & 38.68 / 0.921 / 8.118 & MP-HSIR & 42.16 / 0.968 / 3.030 & 41.17 / 0.965 / 3.242 & 25.89 / 0.618 / 38.911 \\
& Ours & \textbf{29.41} / \textbf{0.797} / \textbf{13.163} & \textbf{33.62} / \textbf{0.711} / \textbf{10.982}  & \textbf{38.91} / \textbf{0.933} / \textbf{6.788} & Ours & \textbf{42.86} / \textbf{0.974} / \textbf{2.229} & \textbf{41.47} / 0.967 / \textbf{2.668} & \textbf{26.11} / \textbf{0.634} / \textbf{28.165}\\

\hline

\multirow{3}{*}{\textbf{Type}} &
\multirow{3}{*}{\textbf{Method}} &
\multicolumn{3}{c|}{\textbf{Completion (Rate = 0.1, 0.2, 0.3)}} &
\multirow{3}{*}{\textbf{Method}} &
\multicolumn{3}{c}{\textbf{Average Results}} \\
\cline{3-5} \cline{7-9}
& & PaviaU & ARAD & Xiong’an & & ICVL & ARAD & RS Data \\
\cline{3-5} \cline{7-9}
& & PSNR / SSIM / SAM & PSNR / SSIM / SAM & PSNR / SSIM / SAM & &PSNR / SSIM / SAM &PSNR / SSIM / SAM &PSNR / SSIM / SAM\\
\hline
\multirow{6}{*}{All in One} 
& InstructIR & 41.75 / 0.803 / 27.723 & 51.91 / 0.946 / 9.761 & 52.50 / 0.830 / 23.305 & InstructIR & 45.28 / 0.954 / 5.224 & 43.30 / 0.934 / 7.844 & 35.94 / 0.666 / 22.765 \\
& PromptIR & 46.77 / 0.889 / 12.177 & 54.79 / 0.999 / 1.020 & 46.37 / 0.931 / 12.714 & PromptIR & 48.69 / 0.984 / 1.281 & 47.20 / 0.984 / 1.510 & 38.19 / 0.812 / 13.247 \\
& DFPIR & 49.97 / 0.801 / 27.994 & 57.63 / 0.884 / 7.654 & 56.55 / 0.831 / 23.474 & DFPIR & 49.41 / 0.962 / 2.732 & 46.43 / 0.956 / 5.838 & 37.44 / 0.679 / 20.437 \\
& MoCE-IR & 46.53 / 0.857 / 20.137 & 57.72 / 0.999 / 0.525 & 47.27 / 0.929 / 17.760 & MoCE-IR & 51.40 / 0.990 / 0.962 & 48.72 / 0.985 / 1.203 & 36.78 / 0.774 / 15.094 \\
& MP-HSIR & 46.76 / 0.878 / \textbf{11.066} & 56.48 / 0.999 / 0.974 & 42.11 / 0.921 / 11.192 & MP-HSIR & 50.69 / 0.989 / 0.978 & 47.85 / 0.984 / 1.608 & 38.33 / 0.839 / 12.734 \\
& Ours & \textbf{50.85} / \textbf{0.901} / 12.862 & \textbf{63.49} / \textbf{1.000} / \textbf{0.422} & \textbf{56.61} / \textbf{0.932} / \textbf{10.281} & Ours & \textbf{51.97} / \textbf{0.990} / \textbf{0.952} & \textbf{51.43} / \textbf{0.989} / \textbf{0.936} & \textbf{39.42} / \textbf{0.851} / \textbf{10.113} \\

\hline
\end{tabular}
\end{adjustbox}
\end{table*}

\section{Experiments}

\subsection{Experimental Settings} 

\noindent \textbf{Datasets.}
We employ eight hyperspectral datasets for training and evaluation, including three natural-scene HSI datasets and five remote sensing HSI datasets.
For natural-scene data, we use ARAD~\cite{arad2022NITRE} and ICVL~\cite{arad2016ICVL} for unified HSI restoration training and testing. Following the protocol in~\cite{wu2025MP-HSIR}, we crop 1,000 images into patches of size \(128 \times 128 \times 31\) for training and reserve 100 non-overlapping images for testing. Additionally, we conduct zero-shot generalization experiments on the CAVE~\cite{CAVE} dataset.
For remote sensing data, we adopt Xiong'an~\cite{yi2020xiongan}, Chikusei~\cite{yokoya2016Chikusei},  PaviaC~\cite{huang2009PaviaU}, PaviaU~\cite{huang2009PaviaU}, and HyRank~\cite{karantzalos2018hyrank}. Specifically, 80\% of each image is randomly cropped and split into patches of size \(128 \!\times\! 128 \!\times\! 100\) for fine-tuning, while the remaining 20\% is used for testing.
Considering the significant domain gap between natural-scene and remote sensing hyperspectral data, and following the established protocol~\cite{wu2025MP-HSIR}, we train separate models for each domain. Detailed descriptions of all datasets are provided in Section B of the supplementary material.

\noindent \textbf{Evaluation Protocols.}
To validate the effectiveness and generalization capability of the proposed model, we consider two main evaluation protocols.
(1) \textit{Unified Training and Evaluation}, where a single model is jointly trained on five HSI restoration tasks, namely Gaussian denoising, Gaussian deblurring, super-resolution, image inpainting, and spectral band completion.
(2) \textit{Zero-Shot Generalization}, where the trained model is directly evaluated on unseen degradation types, specifically motion deblurring and Poisson denoising, without any fine-tuning. 
Detailed configurations for each degradation (\eg, noise levels, blur kernels, sampling ratios) are provided in Section C of the supplementary material.

\noindent \textbf{Implementation Details.}
All experiments are conducted using PyTorch on a single NVIDIA GeForce RTX 4090 GPU. We use the AdamW optimizer with \(\beta_1 = 0.9\) and \(\beta_2 = 0.999\), an initial learning rate of \(1 \times 10^{-4}\), and $L_1$ loss as the objective function, and a batch size of 4.
The model is trained for 3,000 epochs on natural-scene HSIs and 1,500 epochs on remote sensing HSIs.

\noindent \textbf{Evaluation Metrics.}
To quantitatively assess the quality of image restoration, we employ three widely used metrics: Peak Signal-to-Noise Ratio (PSNR) and Structural Similarity Index (SSIM)~\cite{SSIM} for evaluating spatial fidelity, and Spectral Angle Mapper (SAM)~\cite{SAM} for assessing spectral accuracy.

\noindent \textbf{Comparison Methods.}
The comparison methods include 5 unified image restoration methods (PromptIR~\cite{potlapalli2023PromptIR}, InstructIR~\cite{conde2024InstructIR}, DFPIR~\cite{tian2025DFPIR}, MoCE-IR~\cite{zamfir2025MoCEIR}, and MP-HSIR~\cite{wu2025MP-HSIR})  and 12 task-specific methods (Stripformer~\cite{tsai2022stripformer}, DeepRFT~\cite{mao2023DeepRFT}, Loformer~\cite{mao2024Loformer}, ESSA~\cite{ESSAformer}, SRDNet~\cite{liu2024SRD}, VolFormer~\cite{VolFormer}, SSDL~\cite{li2024SSDL}, Restormer~\cite{zamir2022restormer}, PFGIN~\cite{wu2025PFGIN}, SERT~\cite{li2023SERT}, SST~\cite{li2023SST}, and HCANet~\cite{hu2024HCANet}).

\begin{figure*}[tbp]
    
    \includegraphics[width=\linewidth]{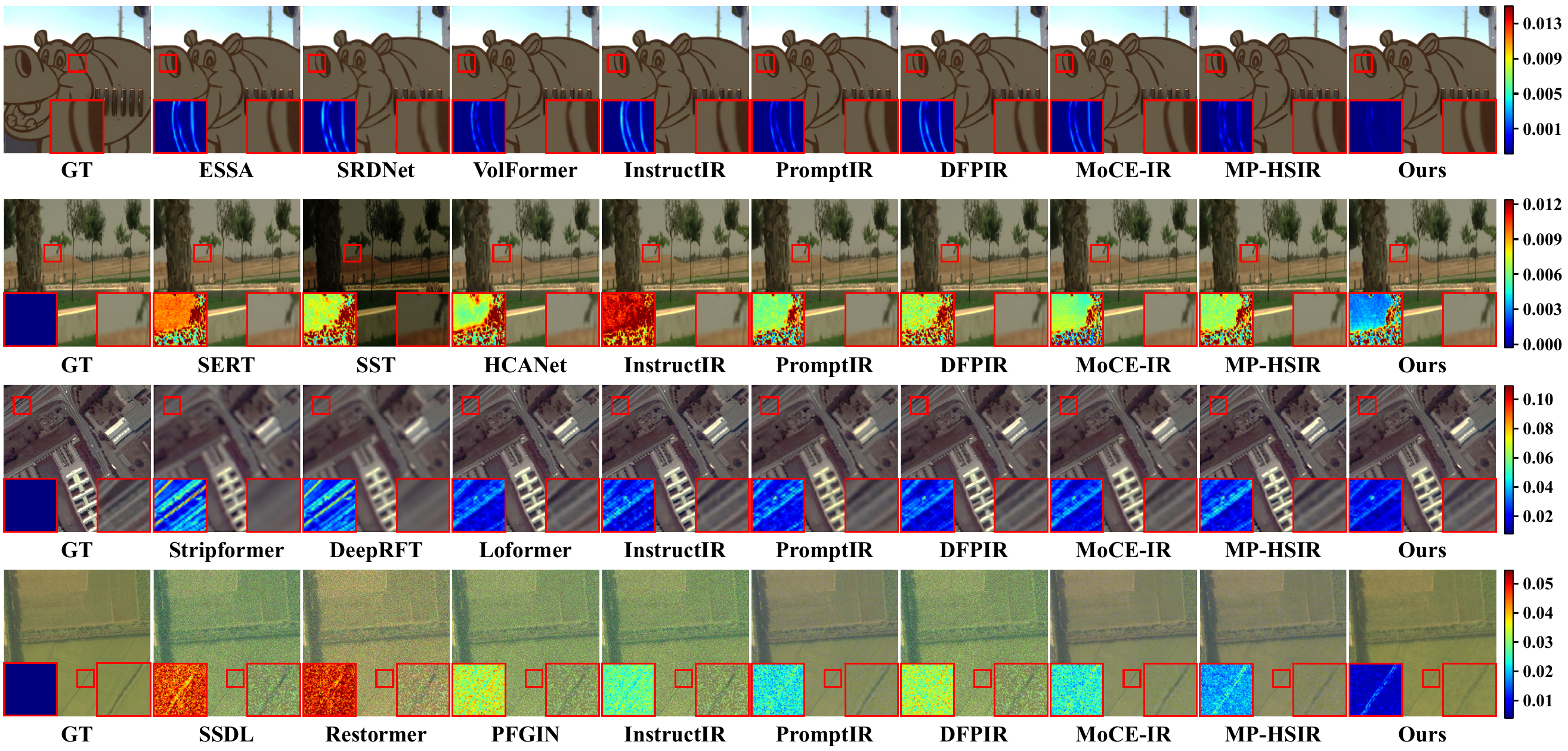}
    \caption{Visual comparison of HSI restoration performance on tasks with known degradation types. From top to bottom: super-resolution on the ARAD dataset~\cite{arad2022NITRE}, denoising on the ICVL dataset~\cite{arad2016ICVL}, deblurring on the PaviaU dataset~\cite{huang2009PaviaU}, and inpainting on the Xiong’an dataset~\cite{yi2020xiongan}. The content within the small red boxes in each image is magnified, with the left side showing the error map compared to the ground truth (GT), and the right side displaying the magnified result images.}
    \label{fig:seen_cmp}
\end{figure*}

\subsection{Main Results}
\noindent \textbf{Quantitative Results on Known Degradation Tasks.} 
We present quantitative comparisons on five degradation tasks in Table~\ref{tab:seen_comparison}. For Gaussian deblurring, Gaussian denoising, inpainting, and super-resolution, we compare both task-specific methods and unified restoration approaches. Due to the scarcity of methods specifically designed for spectral band completion, we only compare against unified restoration baselines for this task.
To ensure a comprehensive evaluation, we randomly evaluate each degradation type across three datasets and further assess average restoration performance separately on natural and remote sensing data. As shown in Table~\ref{tab:seen_comparison}, our method consistently outperforms existing approaches across nearly all metrics for the five degradation types. Moreover, it demonstrates a clear advantage in average performance, validating its effectiveness.

\noindent \textbf{Zero-Shot Generalization to Unknown Degradations.}
To evaluate the zero-shot generalization capability of the proposed DAMP on unseen restoration tasks, we test it on motion deblurring and Poisson noise removal using the CAVE dataset, without any task-specific fine-tuning or explicit knowledge of the degradation process (\eg, blur kernels or noise parameters). As shown in Table~\ref{tab:unseen_cmp}, DAMP achieves superior performance across both tasks, attaining $31.05 dB$ PSNR and $0.899$ SSIM for motion deblurring, outperforming the strongest baseline (PromptIR) by $0.52 dB$, and $24.08 dB$ PSNR for Poisson denoising, a $2.10 dB$ improvement over the previous best method. This superior generalization stems from the fact that DP captures the intrinsic statistical deviations caused by degradation, allowing the router to verify the degradation severity even without seeing the specific degradation type during training.

\begin{table}[htbp]
    
    \caption{Zero-shot performance on unseen tasks.}
    \begin{adjustbox}{width=\linewidth}
    \begin{tabular}{ccc}
        \hline
        \textbf{Method} & \textbf{Motion Deblurring} & \textbf{Poisson Denoising} \\
        \hline
        InstructIR~\cite{conde2024InstructIR} & 28.80 / 0.869 & 18.44 / 0.357 \\
        PromptIR~\cite{potlapalli2023PromptIR} & 30.53 / 0.881 & 21.98 / 0.442 \\
        DFPIR~\cite{tian2025DFPIR} & 30.17 / 0.859 & 18.41 / 0.228 \\
        MoCE-IR~\cite{zamfir2025MoCEIR} & 30.34 / 0.878 & 19.51 / 0.401 \\
        MP-HSIR~\cite{wu2025MP-HSIR} & 23.63 / 0.688 & 16.96 / 0.240 \\
        \textbf{Ours} & \textbf{31.05} / \textbf{0.899} & \textbf{24.08} / \textbf{0.538} \\
        \hline
    \end{tabular}
    \end{adjustbox}
    \label{tab:unseen_cmp}
\end{table}

\begin{figure}[tbp]
    
    \includegraphics[width=\linewidth]{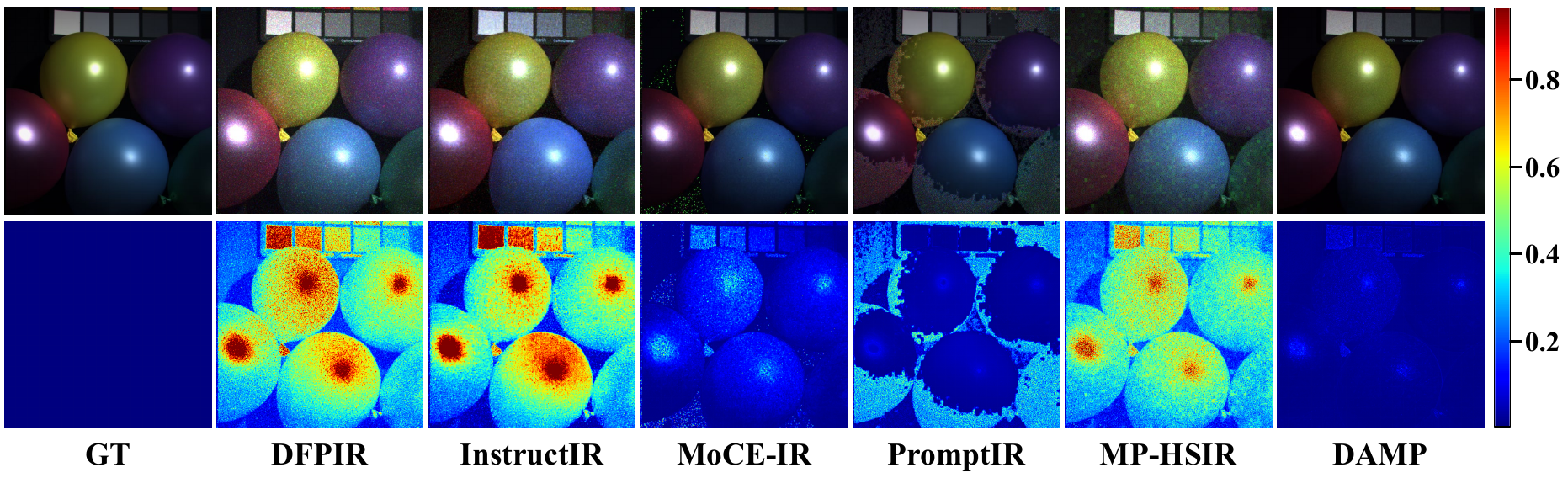}
    \caption{Visual comparison of Poisson denoising on the CAVE dataset~\cite{CAVE}. The first row shows the restoration results, and the second row displays the error maps.}
    \label{fig:unseen_cmp}
\end{figure}

\begin{figure}[tbp]
    
    \includegraphics[width=\linewidth]{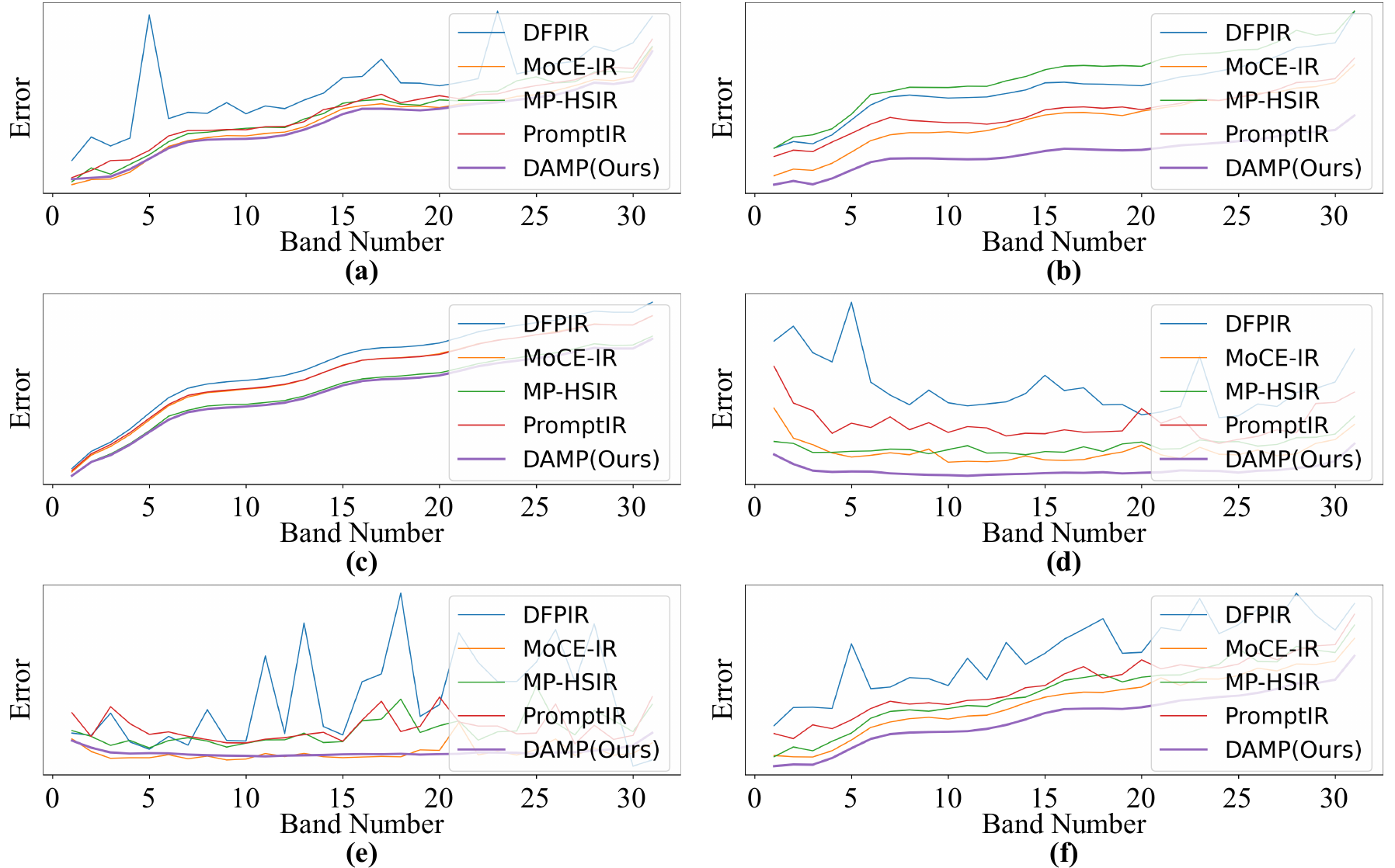}
    \caption{Normalized digital number error across spectral bands on natural datasets for various tasks: (a) Gaussian denoising, (b) Gaussian deblurring, (c) super-resolution, (d) image inpainting, (e) band completion, and (f) average.}
    \label{fig:spec_cmp}
\end{figure}

\noindent \textbf{Visual and Spectral Qualitative Results.}
Fig.~\ref{fig:seen_cmp} presents a visual and error map comparison between our method and competing approaches on four restoration tasks: super-resolution, denoising, deblurring, and inpainting. Our method achieves the lowest reconstruction errors and the highest visual quality across all cases, demonstrating the effectiveness of our degradation-aware prompting strategy.
Fig.~\ref{fig:unseen_cmp} shows qualitative results for HSI restoration under an unseen degradation type, Poisson noise, on the CAVE dataset which was not used during training. Our proposed DAMP produces significantly lower errors than all baseline methods, highlighting its strong generalization capability to previously unseen restoration tasks.
Fig.~\ref{fig:spec_cmp} shows the normalized digital number error between the restored and ground truth images across spectral bands, indicating that our proposed SSAM effectively balances the use of spatial and spectral information and better preserves the intrinsic spectral characteristics of HSIs.

\subsection{Ablation Studies and Analysis}
To validate the effectiveness of the proposed method, we perform unified training for five tasks, namely Gaussian denoising, Gaussian deblurring, super-resolution, image inpainting, and spectral band completion, using natural HSI datasets (ARAD and ICVL). We quantitatively evaluate performance using the average PSNR and SSIM on ARAD.

\noindent \textbf{Effectiveness of Key Components.}
We start with a baseline model that excludes both DP and SSAM. We then progressively incorporate DP and SSAM into the pipeline. As shown in Table~\ref{tab:ablation}, integrating DP significantly boosts the average PSNR by $4.20 dB$ and SSIM by $0.010$, confirming its capability to model diverse degradations. Further adding SSAM yields an additional PSNR gain of $1.41 dB$ and a slight SSIM improvement of $0.003$, demonstrating that the SSAM modulation strategy enhances restoration fidelity. The full model (DP+SSAM) achieves the best performance at $51.43 dB$ PSNR and $0.989$ SSIM.

\begin{table}[htbp]

\caption{Effectiveness of key components.}
\label{tab:ablation}
\begin{adjustbox}{width=\linewidth}
\begin{tabular}{ccccc}
\toprule
 \textbf{DP} & \textbf{SSAM} & \textbf{Average PSNR (dB)↑} & \textbf{Average SSIM↑} \\
\midrule
\ding{55} & \ding{55} & 45.82 & 0.976 \\
\ding{51} & \ding{55} & 50.02 & 0.986 \\
\ding{51} & \ding{51} & \textbf{51.43} & \textbf{0.989} \\
\bottomrule
\end{tabular}
\end{adjustbox}
\end{table}

\noindent \textbf{Impact of Routing Strategies.}
We replace the routing mechanism in DAMP with frequency-based routing~\cite{zamfir2025MoCEIR}, degradation-type routing, and implicit prompt~\cite{wang2025moerl}, respectively, and evaluate their performance. As shown in Table~\ref{tab:routing_comparison}, our DP-based routing  outperforms frequency-based routing by $3.71 dB / 0.006$, degradation-type routing by $5.16 dB / 0.007$, and implicit prompt by$4.62 dB / 0.007$. These results highlight the effectiveness of using DP as a routing function.

\begin{table}[htbp]
\caption{Routing strategy comparison.}
\label{tab:routing_comparison}
\begin{adjustbox}{width=\linewidth}
\begin{tabular}{lcc}
\toprule
\textbf{Routing Strategy} & \textbf{Average PSNR (dB)↑} & \textbf{Average SSIM↑} \\
\midrule
Frequency-based & 47.72 & 0.983 \\
Degradation Type & 46.27 & 0.982 \\
Implicit Prompt & 46.81 & 0.982 \\
\textbf{DP (Ours)} & \textbf{51.43} & \textbf{0.989} \\
\bottomrule
\end{tabular}
\end{adjustbox}
\end{table}

\begin{table}[htbp]
\begin{center}
\caption{Efficiency comparison.}
\label{tab:efficiency}
\begin{adjustbox}{width=0.9\linewidth}
\begin{tabular}{lccc}
\toprule
\textbf{Method} & \textbf{FLOPs (G)} & \textbf{Params (M)} & \textbf{Time (ms)} \\
\midrule
InstructIR~\cite{conde2024InstructIR} & 30.34 & 17.23 & 28.84 \\
PromptIR~\cite{potlapalli2023PromptIR}  & 573.47 & 26.13 & 207.10 \\
DFPIR~\cite{tian2025DFPIR} & 613.80 & 31.33 & 227.22 \\
MoCE-IR~\cite{zamfir2025MoCEIR} & 365.57 & 23.67 & 191.81 \\
MP-HSIR~\cite{wu2025MP-HSIR} & 894.61 & 13.88 & 653.15 \\
\midrule
Ours (w/o DP) & 313.65 & 14.90 & 125.82 \\
Ours (Full)  & 313.80 & 14.91 & 133.18 \\
\bottomrule
\end{tabular}
\end{adjustbox}
    
\end{center}
\end{table}

\begin{figure*}[t]
    \centering
    \includegraphics[width=0.86\textwidth]{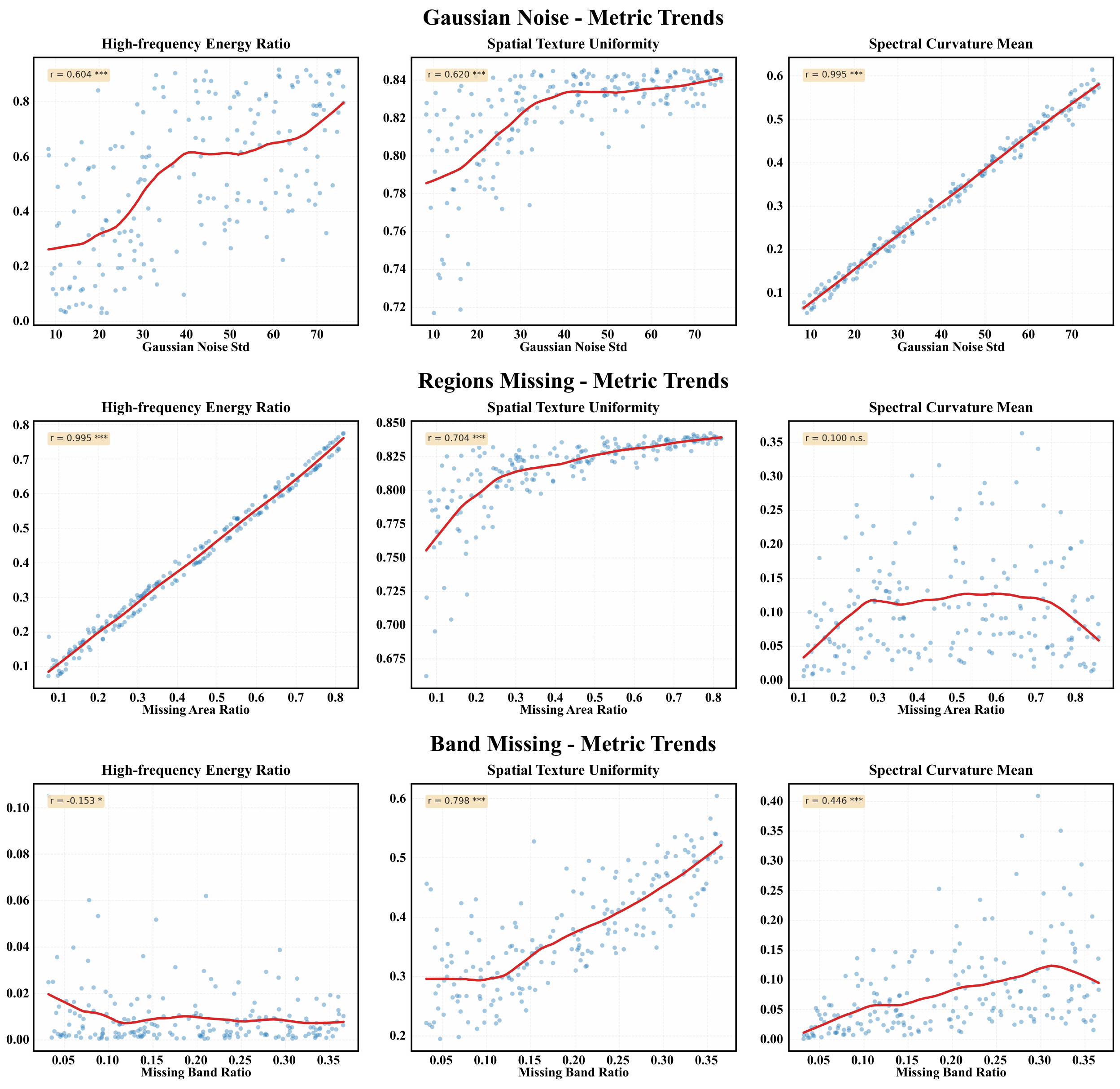}
    \caption{Correlation trends between degradation metrics and degradation severity.}
    \label{fig:metric_severity_trend}
\end{figure*}

\noindent \textbf{Efficiency Analysis.}
Despite its complex adaptive mechanism, DAMP remains computationally efficient. As shown in Table~\ref{tab:efficiency}, DAMP requires fewer FLOPs ($313.8G$) compared to PromptIR ($573.4G$) and MoCE-IR ($365.5G$), while delivering superior performance. The overhead introduced by metric calculation and the DP router is negligible ($<0.2ms$), proving that DAMP is a practical solution for real-world HSI restoration where resources are constrained.

\noindent \textbf{Empirical Validation of Metric-Severity Relationship.}
We conduct quantitative correlation analysis to verify the relation between our proposed degradation metrics and actual degradation severity, which underpins our Degradation Prompt (DP) design. We visualize correlation trends of three continuous degradation types and three representative metrics.
As shown in Fig.~\ref{fig:metric_severity_trend}, most metrics exhibit obvious monotonic trends with increasing degradation severity. Quantitatively, seven of the nine analyzed pairs achieve a Pearson correlation coefficient over 0.4. This evidence demonstrates that integrating multiple complementary metrics enables our method to stably and continuously quantify image degradation severity, facilitating accurate degradation characterization and expert routing.

\section{Conclusion}

In this work, we present DAMP, a unified hyperspectral image restoration framework that replaces impractical explicit degradation labels and opaque black-box embeddings with interpretable Degradation Prompts (DP). By quantifying multi-dimensional degradations via measurable spatial-spectral metrics and using DP as a physical-prior-guided router to orchestrate a Degradation-Adaptive Mixture-of-Experts (DAMoE), DAMP achieves state-of-the-art performance on five standard restoration tasks and exceptional zero-shot generalization to unseen corruptions such as motion blur and Poisson noise. Extensive experiments across natural and remote sensing datasets validate the effectiveness, interpretability, and efficiency of our approach.

\noindent \textbf{Limitations.}
Despite promising results, our method has two remaining limitations. First, while effective on seven tested degradation types, its performance on more diverse real-world HSI corruptions remains unvalidated. Extremely rare or physically distinct degradations may require additional complementary metrics. Second, optimal performance currently requires separate training for natural and remote sensing domains due to inherent domain gaps. A general cross-domain restoration model without retraining remains an open challenge.

\noindent \textbf{Future Work.}
In the future, we will expand our degradation benchmark and develop dynamic metric weighting to handle more diverse corruptions. Moreover, we will explore domain-adaptive techniques to enable a single cross-domain HSI restoration model.

\section*{Acknowledgements.}
This research was supported by the Fundamental and Interdisciplinary Disciplines Breakthrough Plan of the Min-istry of Education of China (No. JYB2025XDXM101), the New Cornerstone Science Foundation through the XPLORER PRIZE, the Innovative Research Group Project of Hubei Province under Grants  2024AFA017, the National Natural Science Foundation of China (624B2109, 62225113, 62331006), and the Zhongguancun Academy Project (20240308).

\section*{Impact Statement}
This paper presents work whose goal is to advance the field of Machine Learning, specifically in the domain of hyperspectral image restoration. The proposed method enhances the quality and utility of hyperspectral data, which has potential positive societal impacts in applications such as environmental monitoring, precision agriculture, and remote sensing. We do not foresee any specific negative societal consequences that must be highlighted here.

\bibliography{main}
\bibliographystyle{icml2026}

\clearpage
\appendix
\section*{Appendix}

This supplementary material provides more details and results that are not included in the main paper due to space limitations. The contents are organized as follows:

\begin{itemize}
\item Section~\ref{detailed metric} presents the detailed metrics selection pipeline.
\item Section~\ref{detailed dataset} provides detailed descriptions of the datasets. 
\item Section~\ref{detail deg} describes the detailed degradation settings.
\item Section~\ref{moer_arch} describes detailed network architecture of DAMP and additional implementation details.
\item Section~\ref{more ablation} presents additional experimental results.
\end{itemize}

\section{Detailed Metric Selection Pipeline}
\label{detailed metric}

To distill a concise yet expressive set of degradation metrics, we design the following metric extraction pipeline.

First, an initial pool of 25 candidate metrics is compiled from the literature in image processing, signal analysis, and remote sensing, focusing on structural integrity, spatial sharpness, and spectral fidelity. These metrics encompass statistical measures derived from four categories:
\begin{itemize}
    \item \textit{Entropy-based}: effective rank, missing data ratio, gradient standard deviation, spatial correlation coefficient, dynamic range compression ratio, normalized consecutive pattern, spectral entropy per pixel, striping artifact index.
    \item \textit{Gradient-based}: mean gradient magnitude, maximum gradient magnitude, high gradient ratio, orientation standard deviation.
    \item \textit{Frequency-based}: high-frequency energy ratio, dominant frequency strength, normalized gap standard deviation, spatial texture uniformity.
    \item \textit{Correlation-based}: maximum curvature, mean adjacent correlation, standard deviation of adjacent correlation, mean correlation, standard deviation of correlation, spatial-spectral consistency deviation, spectral curvature standard deviation, spectral curvature mean, and modulation transfer function approximation.
\end{itemize}

Second, we filter these candidates based on two criteria. The first is \textit{Interpretability}, which requires that each metric exhibits an intuitive and theoretically grounded relationship with specific degradation mechanisms (\eg, blur reduces edge strength). The second is \textit{Modality coverage}, which requires that the selected set jointly captures both spatial and spectral distortions. This filtering step yields a refined subset of 16 viable candidates.

Finally, we conduct empirical validation using a dataset comprising 500 HSI patches from the ARAD dataset~\cite{arad2022NITRE}. Each patch exhibits one of five degradation types, including noise, blur, low resolution, regions missing, or band missing. For each degraded HSI patch, all 16 candidate metrics are computed. Redundant metrics are then removed based on pairwise Pearson correlation coefficients ($\rho < 0.8$), resulting in a final set of 12 non-redundant metrics.

\begin{figure}[tbp]
    \centering
    \includegraphics[width=\linewidth]{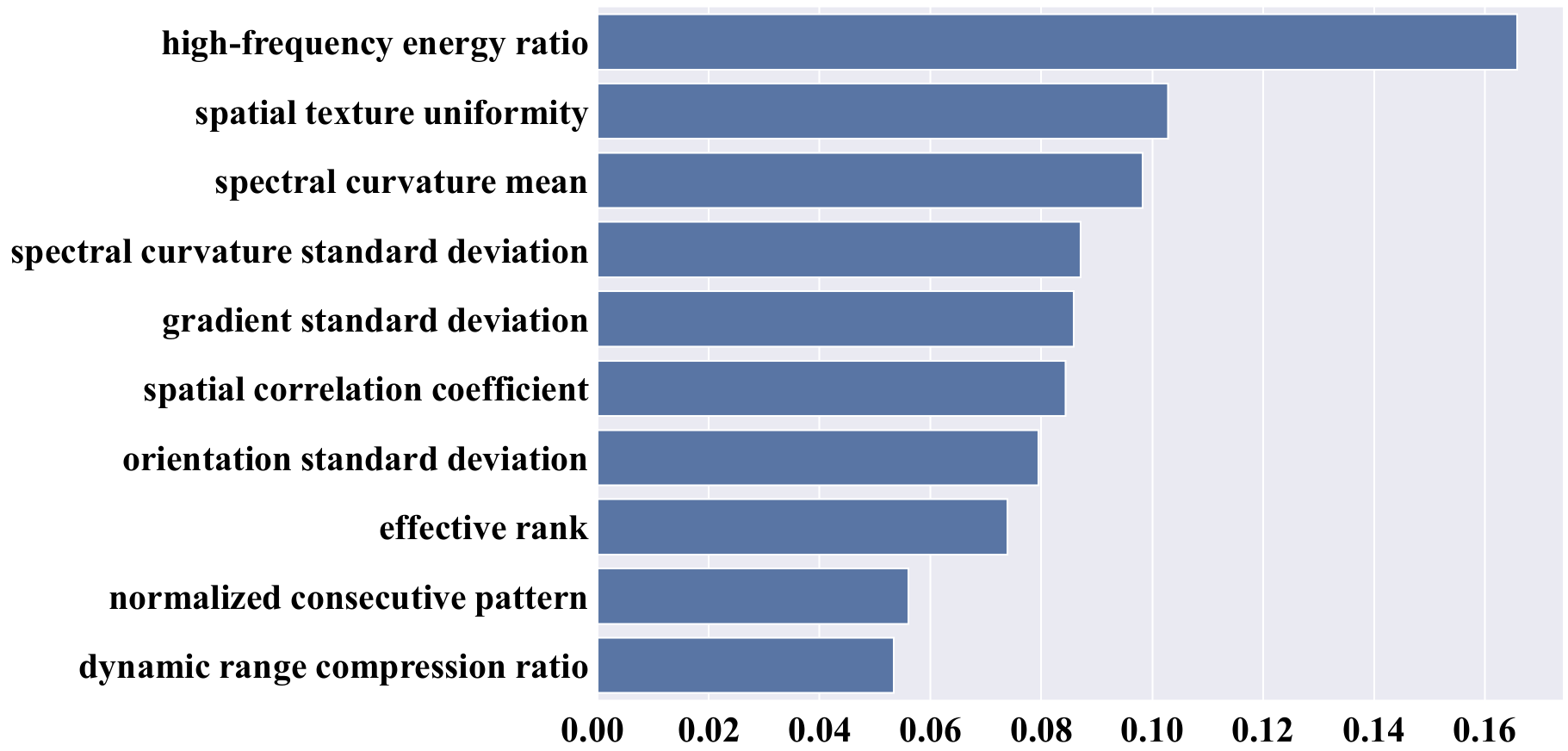}
    \caption{The ten most important metrics for distinguishing degradation types.}
    \label{fig:importance}
\end{figure}

Treating these 12 metrics as input features, a Random Forest classifier is trained (with a 7:3 train--test split) to predict the type of degradation. Feature importance scores from the classifier are used to assess the discriminative power of each metric. The top-10 most important metrics, ranked by their importance scores, are visualized in Fig.~\ref{fig:importance}.

Ultimately, the six most important metrics are selected: \textit{high-frequency energy ratio (HFER)}, \textit{spatial texture uniformity (STU)}, \textit{spectral curvature mean (SCM)}, \textit{spectral curvature standard deviation (SCSD)}, \textit{gradient standard deviation (GSD)}, and \textit{spatial correlation coefficient (SCC)} to form our Degradation Prompts. Their mathematical formulations and physical interpretations are detailed in Table~\ref{tab:metrics_supp}.

\begin{table*}[t]
\centering
\caption{Definitions and physical interpretations of spatial and spectral degradation metrics.}
\label{tab:metrics_supp}
\renewcommand{\arraystretch}{1.8} 
\begin{adjustbox}{max width=\textwidth, center}
\begin{tabular}{l l l}
\toprule
\textbf{Metric} & \textbf{Mathematical Formulation} & \textbf{Physical Interpretation} \\
\midrule
HFER & $ \frac{1}{C} \sum_{c=1}^{C} \frac{\sum_{(u,v) \in \Omega_H} |\mathcal{F}[x_c(u,v)]|^2}{\sum_{(u,v)} |\mathcal{F}[x_c(u,v)]|^2}$ & Proportion of high-frequency energy indicating spatial detail preservation \\
STU & $\frac{1}{C} \sum_{c=1}^{C} \frac{\exp\left(\frac{1}{HW} \sum_{h,w} \ln|\mathcal{F}[x_c(h,w)]|\right)}{\frac{1}{HW} \sum_{h,w} |\mathcal{F}[x_c(h,w)]|}$ & Spectral smoothness measured by geometric-to-arithmetic mean ratio \\
SCM & $ \frac{1}{C-2} \sum_{i=1}^{C-2} |\nabla^2 s_i|$ & Average absolute curvature of spectral curves indicating spectral continuity \\
SCSD & $ \sqrt{\frac{1}{C - 3}\sum_{i=1}^{C - 2}\left(\nabla^2 s_i -\frac{1}{C - 2}\sum_{j=1}^{C - 2}\nabla^2 s_j\right)^2}$ & Standard deviation of spectral curvature reflecting shape irregularity across wavelengths \\
GSD & $\sqrt{\frac{1}{HW - 1}\sum_{i=1}^{HW}\left(M_{b,i}-\frac{1}{HW}\sum_{j=1}^{HW}M_{b,j}\right)^2}$ & Standard deviation of gradient magnitudes indicating spatial heterogeneity of image details \\
SCC & $\rho = \frac{1}{C} \sum_{c=1}^{C} \frac{1}{2} \left( \operatorname{corr}(\mathbf{x}_c, \mathbf{x}_c^{\rightarrow}) + \operatorname{corr}(\mathbf{x}_c, \mathbf{x}_c^{\downarrow}) \right)$ & Spatial correlation coefficient reflecting local pixel redundancy and structural smoothness. \\
\bottomrule
\end{tabular}
\end{adjustbox}
\end{table*}

\begin{table*}[htbp]
\centering
\caption{Detailed descriptions of the datasets.}
\label{tab:dataset}
\begin{adjustbox}{max width=\textwidth, center}
\begin{tabular}{cccccc}
\toprule
\textbf{Dataset} & \textbf{Sensor} & \textbf{Wavelength (nm)} & \textbf{Channels} & \textbf{Size} & \textbf{GSD (m)} \\
\midrule
ARAD~\cite{arad2022NITRE} & Specim IQ & 400--700 & 31 & 482$\times$512 & / \\
\midrule
ICVL~\cite{arad2016ICVL} & Specim PS Kappa DX4 & 400--700 & 31 & 1392$\times$1300 & / \\
\midrule
CAVE~\cite{CAVE} & Specim PFD-1000 VNIR & 400--700 & 31 & 512$\times$512 & / \\
\midrule
Xiong'an~\cite{yi2020xiongan} & Unknown & 400--1000 & 256 & 3750$\times$1580 & 0.5 \\
\midrule
Chikusei~\cite{yokoya2016Chikusei} & HH-VNIR-C & 343--1018 & 128 & 2517$\times$2335 & 2.5 \\
\midrule
PaviaC~\cite{huang2009PaviaU} & ROSIS & 430--860 & 102 & 1096$\times$715 & 1.3 \\
\midrule
PaviaU~\cite{huang2009PaviaU} & ROSIS & 430--860 & 103 & 610$\times$340 & 1.3 \\
\midrule
HyRank~\cite{karantzalos2018hyrank} & EO-1 Hyperion & 400--2500 & 176 & 250$\times$1376 & 30 \\

\bottomrule
\end{tabular}
\end{adjustbox}
\end{table*}

\begin{table*}[htbp]
\centering
\caption{Comparison of single-task and multi-task training across five restoration tasks.}
\label{tab:single_vs_multi}
\begin{adjustbox}{width=0.9\linewidth}
\begin{tabular}{l|c|c|c|c|c}
\toprule
\textbf{Training Mode} 
& \textbf{Gaussian Denoising}
& \textbf{Gaussian Deblurring}
& \textbf{Super-Resolution}
& \textbf{Image Inpainting}
& \textbf{Band Completion}\\
\midrule
Single-Task & 41.06 / 0.964 & \textbf{53.09} / \textbf{0.998} & 43.37 / 0.978 & 53.26 / 0.999 & 62.59 / 1.000 \\
Multi-Task  & \textbf{41.47} / \textbf{0.967} & 52.84 / \textbf{0.998} & \textbf{44.01} / \textbf{0.981} & \textbf{55.35} / \textbf{0.999} & \textbf{63.49} / \textbf{1.000} \\
\bottomrule
\end{tabular}
\end{adjustbox}
\end{table*}

\section{Detailed Descriptions of All Datasets}
\label{detailed dataset}

We employ eight hyperspectral datasets for training and evaluation, comprising three natural HSI datasets, namely ARAD~\cite{arad2022NITRE}, ICVL~\cite{arad2016ICVL}, and CAVE~\cite{CAVE}, and five remote sensing HSI datasets, namely Xiong’an~\cite{yi2020xiongan}, Chikusei~\cite{yokoya2016Chikusei}, PaviaC~\cite{huang2009PaviaU}, PaviaU~\cite{huang2009PaviaU}, and HyRank~\cite{karantzalos2018hyrank}. Details regarding the sensors used, spectral wavelength ranges, number of spectral bands, image dimensions, and spatial resolution for remote sensing datasets are summarized in Table~\ref{tab:dataset}.

\noindent \textbf{ARAD~\cite{arad2022NITRE}.}  
The ARAD dataset was captured using the Specim IQ portable imaging spectrometer. It contains 1,000 images. Following the standard public protocol, we use 900 images for training and 50 for testing.

\noindent \textbf{ICVL~\cite{arad2016ICVL}.}  
The ICVL dataset was acquired with the Specim PS Kappa DX4 imaging spectrometer and consists of 201 images. We adopt the same data split as in~\cite{wu2025MP-HSIR}, using 100 images for training and 50 for testing.

\noindent \textbf{CAVE~\cite{CAVE}.}  
The CAVE dataset is a widely used indoor hyperspectral imaging benchmark collected by the Specim PFD-1000 VNIR system. We employ the entire CAVE dataset for testing.

\noindent \textbf{Xiong'an~\cite{yi2020xiongan}.}  
The Xiong’an dataset comprises high-resolution airborne hyperspectral imagery over China’s Xiong’an New Area, covering a broad spectral range from the visible to the short-wave infrared, with 256 spectral bands. Three $512 \times 512$ regions are randomly cropped for testing, while the rest of the data is used for training.

\noindent \textbf{Chikusei~\cite{yokoya2016Chikusei}.}  
The Chikusei dataset was collected over the Chikusei area in Japan using the HH-VNIR-C sensor, spanning continuous spectral bands from blue to near-infrared with 128 channels. Four $512 \times 512$ regions are randomly selected for testing, and the remaining areas serve as training data.

\noindent \textbf{PaviaC~\cite{huang2009PaviaU}.}  
PaviaC is a classical urban hyperspectral remote sensing dataset acquired over Pavia, Italy, by the ROSIS (Reflective Optics System Imaging Spectrometer) sensor mounted on an airborne platform. It features a rich mix of buildings, roads, and vegetation across 102 spectral bands. A $256 \times 256$ region is randomly cropped for testing, with the remainder used for training.

\noindent \textbf{PaviaU~\cite{huang2009PaviaU}.}  
PaviaU is a subset of PaviaC, focusing on the University of Pavia campus. It was also captured by the ROSIS sensor and contains 103 spectral bands. A $256 \times 256$ region is randomly selected for testing, while the rest is used for training.

\noindent \textbf{HyRank~\cite{karantzalos2018hyrank}.}  
The HyRank dataset is derived from NASA’s EO-1 satellite, which is equipped with the Hyperion sensor. It covers spectral bands from the visible to the short-wave infrared (SWIR) with 176 channels. Two $128 \times 128$ regions are randomly cropped for testing, and the remaining data is used for training.

\section{Detailed Configurations for Degradations}
\label{detail deg}

To comprehensively evaluate the versatility and robustness of our model across diverse hyperspectral image restoration tasks, we consider seven representative degradation scenarios, each designed to reflect real-world challenges in HSI acquisition and processing. These include:

\noindent \textbf{Gaussian Denoising.}  
Each image is corrupted with zero-mean, independent and identically distributed Gaussian noise, where the standard deviation $\sigma$ ranges from 30 to 70. For evaluation, we select three representative noise levels: $\sigma = 30$, $50$, and $70$.

\noindent \textbf{Super-Resolution.}  
Each image is downsampled using bicubic interpolation with scale factors of $2$ and $4$. To maintain consistent input and output dimensions for the all-in-one model, an unpooling operation is applied to upsample the degraded HSIs back to their original spatial resolution.

\noindent \textbf{Gaussian Deblurring.}  
Each image is blurred using an isotropic Gaussian kernel whose standard deviation $\sigma$ is determined from the kernel size $K_S$ via  
\begin{equation}
    \sigma = 0.3 \times \left( \frac{K_S - 1}{2} - 1 \right) + 0.8.
    \label{eq:gaussian_sigma}
\end{equation}
We employ kernel sizes $K_S = 9$ and $15$ for both natural and remote sensing hyperspectral datasets.

\noindent \textbf{Inpainting.}  
Each image is masked with a random binary mask. We consider three severe occlusion levels with missing pixel rates of $0.7$, $0.8$, and $0.9$.

\noindent \textbf{Band Completion.}  
A fixed proportion of spectral bands is randomly removed from each image, with discard rates of $0.1$, $0.2$, and $0.3$.

\noindent \textbf{Motion Deblurring.}  
Each image is degraded using a linear motion blur kernel with a radius of $15$ pixels and an orientation of $45^\circ$. The pre-trained model is directly evaluated without further adaptation.

\noindent \textbf{Poisson Denoising.}  
Each image is corrupted by Poisson noise, where the noise intensity is controlled by a scaling factor of $10$. The pre-trained model is directly evaluated on this degradation setting.

The exact degradation generation formulas are available at our official GitHub repository: \url{https://github.com/MiliLab/DAMP}.

\section{Network Architecture and Additional Implementation Details}
\label{moer_arch}
We have introduced the overall pipeline of DAMP in Section 3.3 of the main paper. This section supplements detailed architectural configurations and experimental settings.

In terms of network design, the base channel number is set to 64. The four-level encoder is stacked with 1, 2, 2, and 3 encoder blocks from top to bottom, while the corresponding three-level decoder consists of 2, 2 and 1 decoder blocks in a bottom-up order. Downsampling modules use $3\times3$ convolutions with stride 2, and upsampling modules adopt $2\times2$ transposed convolutions with stride 2. The refinement module is built with two cascaded Transformer blocks sharing the same structure as encoder blocks. Moreover, in the Degradation-Adaptive MoE module, each degradation-specific expert adopts feature dimensions one quarter of those used by shared experts.

For training strategies, we employ the AdamW optimizer with an initial learning rate of $1\times10^{-4}$, and adopt the CosineAnnealingLR learning rate scheduling scheme. The batch size is set to 4, and random cropping is utilized for online data augmentation during model training.

\begin{figure}[htbp]
    \begin{center}
    \centerline{
    \includegraphics[width=\linewidth]{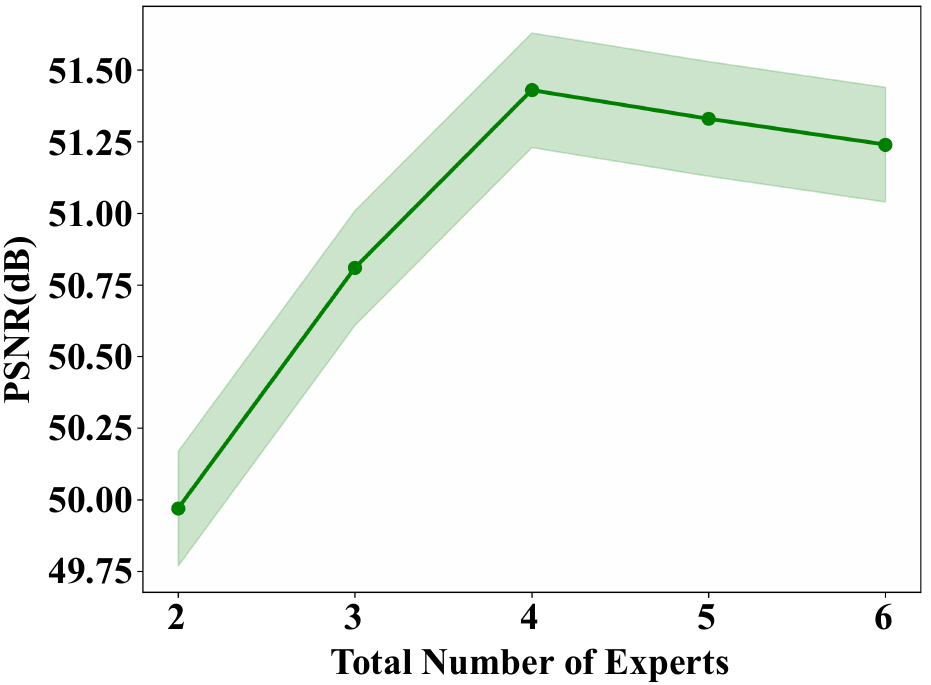}}
    \caption{Impact of total number of experts.}
    \label{fig:expert1}
    \end{center}
\end{figure}

\section{Additional Experimental Results}
\label{more ablation}

\subsection{Multi-Task Collaboration in DAMP}
To evaluate DAMP’s ability to facilitate cross-task synergy, we compare its performance under single-task training versus multi-task collaborative training across five HSI restoration tasks: Gaussian denoising, super-resolution, Gaussian deblurring, image inpainting, and spectral band completion. The models are trained on a natural HSI dataset and evaluated on the ARAD dataset~\cite{arad2022NITRE}.

As shown in Table~\ref{tab:single_vs_multi}, multi-task collaborative training achieves superior performance in four out of the five tasks. Notably, significant gains are observed in image inpainting ($+2.09 dB$) and spectral band completion ($+0.90 dB$), while maintaining comparable SSIM values. These results indicate that DAMP effectively mitigates conflicts among different restoration tasks and leverages shared representations across diverse degradation types to enhance the performance of individual tasks.

\subsection{Ablation Study on the Number of Experts}
We conduct ablation studies on the number of experts in the MoE component of the proposed DAMP framework. The experiments are conducted on the natural HSI dataset and evaluated on the ARAD dataset~\cite{arad2022NITRE}, using the average PSNR and SSIM across five tasks: Gaussian denoising, super-resolution, Gaussian deblurring, image inpainting, and spectral band completion.

\begin{table}[htbp]
\centering
\caption{Ablation on the total number of experts (top-1 routing).}
\label{tab:total_experts}
\begin{adjustbox}{width=\linewidth}
\begin{tabular}{cccc}
\toprule
 \textbf{Total Experts} & \textbf{Average PSNR (dB)↑} & \textbf{Average SSIM↑} \\
\midrule
2 & 49.97 & 0.985 \\
3 & 50.81 & 0.986 \\
4 & \textbf{51.43} & \textbf{0.989} \\
5 & 51.33 & 0.988 \\
6 & 51.24 & 0.987 \\
\bottomrule
\end{tabular}
\end{adjustbox}
\end{table}

\begin{figure}[tbp]
    \begin{center}
    \centerline{\includegraphics[width=\linewidth]{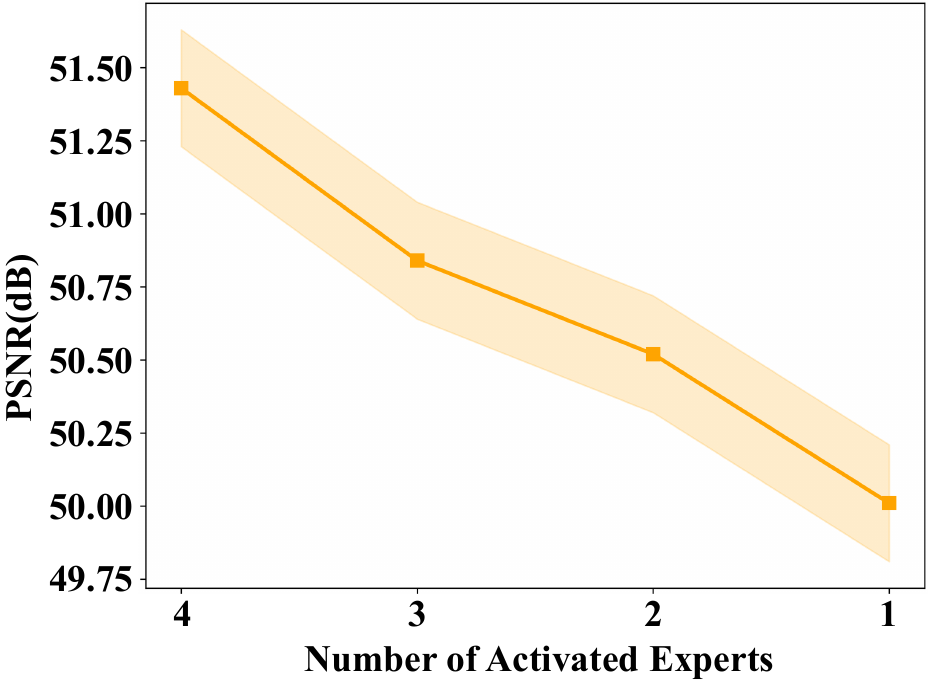}}
    \caption{Impact of number of activated experts.}
    \label{fig:expert2}
    \end{center}
\end{figure}

First, we investigate the effect of the total number of experts while keeping the number of activated experts fixed at 1. As shown in Table~\ref{tab:total_experts} and Fig.~\ref{fig:expert1}, the average PSNR and SSIM first increase and then slightly decrease as the total number of experts grows from 2 to 6. The best performance is achieved with 4 experts, indicating that this configuration provides sufficient capacity to effectively recover hyperspectral images under a wide range of degradation patterns. Beyond four experts, the marginal gains diminish, likely due to increased model complexity without corresponding improvements in representational diversity.

\begin{table}[htbp]
\centering
\caption{Ablation on the number of activated experts (total experts fixed at 4).}
\label{tab:activated_experts}
\begin{adjustbox}{width=\linewidth}
\begin{tabular}{cccc}
\toprule
 \textbf{Activated Experts} & \textbf{Average PSNR (dB)↑} & \textbf{Average SSIM↑} \\
\midrule
1 & \textbf{51.43} & \textbf{0.989} \\
2 & 50.84 & 0.986 \\
3 & 50.52 & 0.986 \\
4 & 50.01 & 0.985 \\
\bottomrule
\end{tabular}
\end{adjustbox}
\end{table}

Next, with the total number of experts fixed at 4, we vary the number of activated experts from 1 to 4. The results, presented in Table~\ref{tab:activated_experts} and Fig.~\ref{fig:expert2}, show that activating only a single expert yields the highest performance. As more experts are activated simultaneously, the overall performance gradually declines. This suggests that specialization among experts is crucial. Each expert learns to handle a specific pattern of degradation, and combining multiple experts may introduce interference or redundancy, thereby reducing reconstruction quality.

We further conduct an ablation study to analyze the sensitivity of our MoE routing mechanism to Gaussian noise. In this experiment, we fix all other hyperparameters and only vary the standard deviation $\sigma$ of the Gaussian noise added to the routing logits. All experiments are performed on the ARAD validation set.

The results in Table~\ref{tab:routing_noise_sensitivity} show that adding an appropriate amount of routing noise can improve model performance, while zero noise or excessive noise leads to performance degradation. Specifically, our model maintains stable performance across a wide range of noise levels ($\sigma$ from 0.25 to 1.5), with the best performance achieved at $\sigma$=1.0. This validates the strong robustness of our routing mechanism against noise perturbations.

\begin{table}[htbp]
    \centering
    \caption{Routing noise sensitivity analysis.}
    \label{tab:routing_noise_sensitivity}
    \begin{adjustbox}{width=0.8\linewidth}
    \begin{tabular}{c c c c}
    \toprule
    Noise Std ($\sigma$) & PSNR (dB)↑ & SSIM↑ & SAM↓ \\
    \midrule
    0 & 50.42 & 0.985 & 1.142 \\
    0.25 & 51.38 & 0.989 & 0.957 \\
    1.0 & \textbf{51.43} & \textbf{0.989} & \textbf{0.936} \\
    1.5 & 51.37 & 0.988 & 1.052 \\
    3.0 & 50.87 & 0.986 & 1.073 \\
    \bottomrule
    \end{tabular}
    \end{adjustbox}
\end{table}

\subsection{Handling Spatially Heterogeneous/Localized Degradations}
\label{sec:appendix_local_degradations}

We conduct an additional analysis experiment to verify the effectiveness of our Degradation Prompt (DP) in handling spatially heterogeneous and localized degradations, which are prevalent in real-world hyperspectral imaging scenarios. The DP naturally handles such degradations through its joint 6-dimensional spatial-spectral representation. Localized degradations uniquely alter global statistics (e.g., regional missing pixels simultaneously lower HFER and increase SCM due to spatial-spectral discontinuities).

\begin{figure}[t]
    \centering
    \includegraphics[width=\linewidth]{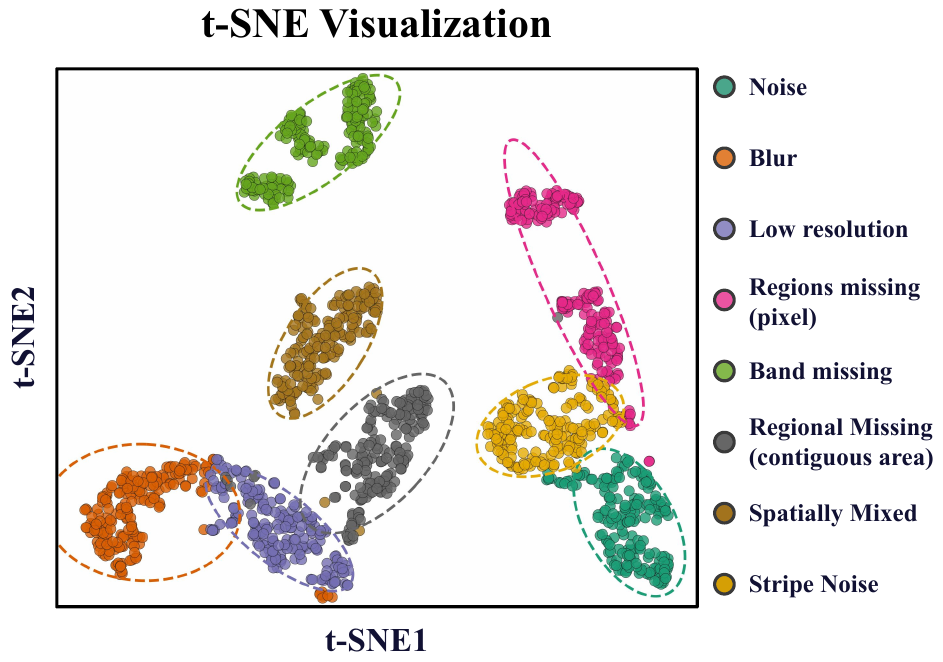}
    \caption{T-SNE clustering based on DPs. The clustering results show that the DPs can effectively handle spatially heterogeneous corruptions.}
    \label{fig:tsne_local_degradations}
\end{figure}

To validate this, we simulate three spatially heterogeneous scenarios on the ARAD dataset: (1) \textbf{Regional missing} (contiguous blocks covering 10\%-30\% of the image area); (2) \textbf{Spatially mixed degradation} (global noise + local blur); and (3) \textbf{Stripe noise} (20\% of horizontal/vertical stripes interrupted). Using the standard global DP and routing, DAMP effectively characterizes these mixed states and outperformed SOTA models. 
Fig.~\ref{fig:tsne_local_degradations} presents the t-SNE visualizations of DP embeddings for these localized degradations, confirming they form distinct and well-separated clusters.

\begin{table}[t]
    \centering
    \caption{Performance under spatially heterogeneous and mixed degradations (ARAD dataset).}
    \label{tab:local_degradations}
    \resizebox{\linewidth}{!}{
    \begin{tabular}{l l c c c}
    \toprule
    Degradation Scenario & Method & PSNR (dB) & SSIM & SAM \\
    \midrule
    \multirow{3}{*}{Regional Missing } & PromptIR & 35.21 & 0.915 & 3.459 \\
    & MP-HSIR & 35.76 & 0.918 & 3.213 \\
    & \textbf{DAMP (Ours)} & \textbf{36.43} & \textbf{0.933} & \textbf{2.896} \\
    \midrule
    \multirow{3}{*}{Spatially Mixed} & PromptIR & 40.86 & 0.968 & 1.938 \\
    & MP-HSIR & 39.47 & 0.962 & 2.213 \\
    & \textbf{DAMP (Ours)} & \textbf{42.38} & \textbf{0.974} & \textbf{1.768} \\
    \midrule
    \multirow{3}{*}{Stripe Noise} & PromptIR & 44.56 & 0.982 & 1.997 \\
    & MP-HSIR & 45.73 & 0.983 & 1.733 \\
    & \textbf{DAMP (Ours)} & \textbf{47.85} & \textbf{0.985} & \textbf{1.214} \\
    \bottomrule
    \end{tabular}
    }
\end{table}

\subsection{Performance under Extreme Combined Degradations}
\label{sec:appendix_extreme_degradations}

While relying on a single metric (e.g., SCM) could lead to brittleness under extreme non-linear degradations, DAMP mitigates this via its \textbf{joint 6-dimensional spatial-spectral representation}. Even if complex degradations mask classical high-frequency or spectral shifts, they uniquely alter the \textit{joint distribution} of these six metrics, ensuring a highly resilient and discriminative signature.

To empirically validate this, we stress-test DAMP under an extreme combined degradation scenario: \textbf{Gaussian Noise + Band Missing + Motion Blur + Regions Missing}. As shown in Table~\ref{tab:extreme_degradations}, despite the severe corruption of standard image statistics, DAMP maintains robust performance and significantly outperforms SOTA unified models.

\begin{table}[t]
    \centering
    \caption{Performance under extreme combined degradation.}
    \label{tab:extreme_degradations}
    \resizebox{0.8\linewidth}{!}{
    \begin{tabular}{l c c c}
    \toprule
    Method & PSNR (dB) & SSIM & SAM \\
    \midrule
    PromptIR & 33.89 & 0.883 & 8.847 \\
    MP-HSIR & 33.42 & 0.864 & 9.392 \\
    \textbf{DAMP (Ours)} & \textbf{35.82} & \textbf{0.903} & \textbf{6.531} \\
    \bottomrule
    \end{tabular}
    }
\end{table}

\subsection{Zero-Shot Comparisons with Task-Specific Fully-Supervised SOTAs}
\label{sec:appendix_zero_shot_task_sota}

We conduct additional zero-shot comparisons against fully-supervised, task-specific SOTA models to further validate the generalization ability of our DAMP framework. Specifically, we compare with MLWNet~\cite{gao2024efficient} for motion deblurring and LDERT~\cite{li2024hsidiff} for Poisson denoising on the CAVE dataset. To ensure fair comparison, we split the CAVE dataset into 20 training images and 12 testing images, strictly adhering to the training and evaluation protocols of the baseline models.

A key distinction lies in the model deployment paradigm: DAMP operates as a single unified model in a zero-shot manner, requiring no task-specific fine-tuning, explicit degradation priors, or kernel information. In contrast, the compared SOTA models are custom-designed architectures trained in a fully supervised manner exclusively on their respective target degradation tasks. Despite this inherent disadvantage, DAMP achieves competitive zero-shot performance. Furthermore, when fine-tuned with only 20\% (4 images) of the target training data, DAMP surpasses the performance of these task-specific fully-supervised SOTA models.

\begin{table}[t]
    \centering
    \caption{Zero-shot vs. task-specific SOTA comparison (CAVE dataset).}
    \label{tab:zero_shot_task_sota}
    \resizebox{\linewidth}{!}{
    \begin{tabular}{l l l c c c}
    \toprule
    Task & Method & Model Type & PSNR (dB) & SSIM & SAM \\
    \midrule
    \multirow{3}{*}{Motion Deblurring} & MLWNet & Fully Supervised & 33.74 & 0.923 & 13.257 \\
    & DAMP & Zero-Shot & 32.57 & 0.915 & 15.018 \\
    & DAMP & Finetuned & 36.56 & 0.937 & 10.843 \\
    \midrule
    \multirow{3}{*}{Poisson Denoising} & LDERT & Fully Supervised & 26.48 & 0.723 & 18.359 \\
    & DAMP & Zero-Shot & 25.12 & 0.638 & 20.136 \\
    & DAMP & Finetuned & 28.86 & 0.756 & 15.124 \\
    \bottomrule
    \end{tabular}
    }
\end{table}

\subsection{Comparisons with Unsupervised and Self-Supervised SOTA Models}
\label{sec:appendix_unsupervised_self_supervised}

We conduct comprehensive comparisons against state-of-the-art unsupervised and self-supervised models to further demonstrate the superiority of our DAMP framework. Experiments are performed on both the Natural (ARAD) and Remote Sensing (RS) datasets. We compare with two unsupervised models: DDS2M~\cite{miao2023dds2m} and HIR-Diff~\cite{pang2024hsidiff}, as well as one self-supervised model: TBSN~\cite{li2025rethinking}.

As shown in Table~\ref{tab:unsupervised_self_supervised}, DAMP significantly outperforms all these baselines. While unsupervised and self-supervised methods eliminate the need for clean ground-truth data during training, our fully-supervised unified architecture with interpretable degradation prompting achieves substantially higher spatial restoration fidelity (PSNR/SSIM) and spectral accuracy (SAM) across both datasets.

\begin{table}[htbp]
    \centering
    \caption{Comparison with unsupervised and self-supervised SOTA models.}
    \label{tab:unsupervised_self_supervised}
    \begin{adjustbox}{width=\linewidth}
    \begin{tabular}{l c c}
    \toprule
    Method & ARAD (PSNR / SSIM / SAM) & RS (PSNR / SSIM / SAM) \\
    \midrule
    DDS2M & 40.11 / 0.916 / 7.182 & 32.13 / 0.602 / 25.354 \\
    HIR-Diff & 41.22 / 0.921 / 6.124 & 33.22 / 0.623 / 22.156 \\
    TBSN & 42.16 / 0.924 / 6.553 & 32.35 / 0.647 / 22.446 \\
    \textbf{DAMP (Ours)} & \textbf{51.43 / 0.989 / 0.936} & \textbf{39.42 / 0.851 / 10.113} \\
    \bottomrule
    \end{tabular}
    \end{adjustbox}
\end{table}

\subsection{Ablation Study on Degradation Prompt Metrics}
\label{sec:appendix_ablation_dp_metrics}

We conduct a leave-one-out ablation study to verify the individual contribution of each metric in our 6-dimensional Degradation Prompt (DP) representation. Experimental results show that removing any single metric leads to degraded restoration performance, especially in terms of zero-shot generalization ability.

As presented in Table~\ref{tab:ablation_dp_leave_one_out}, removing the top spatial metric (HFER) or spectral metric (SCM) causes the most significant performance drop across all evaluation indicators. This confirms that each dimension of our joint 6-metric representation is indispensable for accurately characterizing complex and diverse degradation patterns.

\begin{table}[htbp]
    \centering
    \caption{Leave-one-out ablation for degradation prompts.}
    \label{tab:ablation_dp_leave_one_out}
    \begin{adjustbox}{width=0.85\linewidth}
    \begin{tabular}{l c c c}
    \toprule
    Removed Metric & PSNR (dB)↑ & SSIM↑ & SAM↓ \\
    \midrule
    None & \textbf{51.43} & \textbf{0.989} & \textbf{0.936} \\
    w/o HFER & 50.18 & 0.981 & 1.245 \\
    w/o SCM & 50.25 & 0.982 & 1.384 \\
    w/o STU & 50.42 & 0.983 & 1.182 \\
    w/o SCSD & 50.76 & 0.985 & 1.121 \\
    w/o GSD & 50.91 & 0.986 & 1.056 \\
    w/o SCC & 51.12 & 0.987 & 0.992 \\
    \bottomrule
    \end{tabular}
    \end{adjustbox}
\end{table}

We further compare the performance of our final 6-metric set with the initial 3-metric and full 12-metric candidate sets to validate the effectiveness of our metric selection pipeline. The results in Table~\ref{tab:ablation_metric_combinations} show that our final 6-metric configuration outperforms the initial 3-metric set by 1.41 dB in PSNR while achieving nearly identical performance to the 12-metric set. This demonstrates that our selection strategy effectively balances model computational efficiency and degradation representational power.

\begin{table}[htbp]
    \centering
    \caption{Ablation on metric combinations.}
    \label{tab:ablation_metric_combinations}
    \begin{adjustbox}{width=0.9\linewidth}
    \begin{tabular}{l c c c}
    \toprule
    Metric Combination & Average PSNR (dB)↑ & Average SSIM↑ & Average SAM↓ \\
    \midrule
    Initial 3 Metrics & 50.02 & 0.980 & 1.412 \\
    Final 6 Metrics (Ours) & \textbf{51.43} & \textbf{0.989} & \textbf{0.936} \\
    12 Candidate Metrics & 51.51 & 0.989 & 0.935 \\
    \bottomrule
    \end{tabular}
    \end{adjustbox}
\end{table}

\subsection{Visualization and Quantification of Spatial/Spectral Experts}
\label{sec:appendix_expert_specialization_visualization}

\begin{figure*}[htbp]
    \centering
    \includegraphics[width=0.9\linewidth]{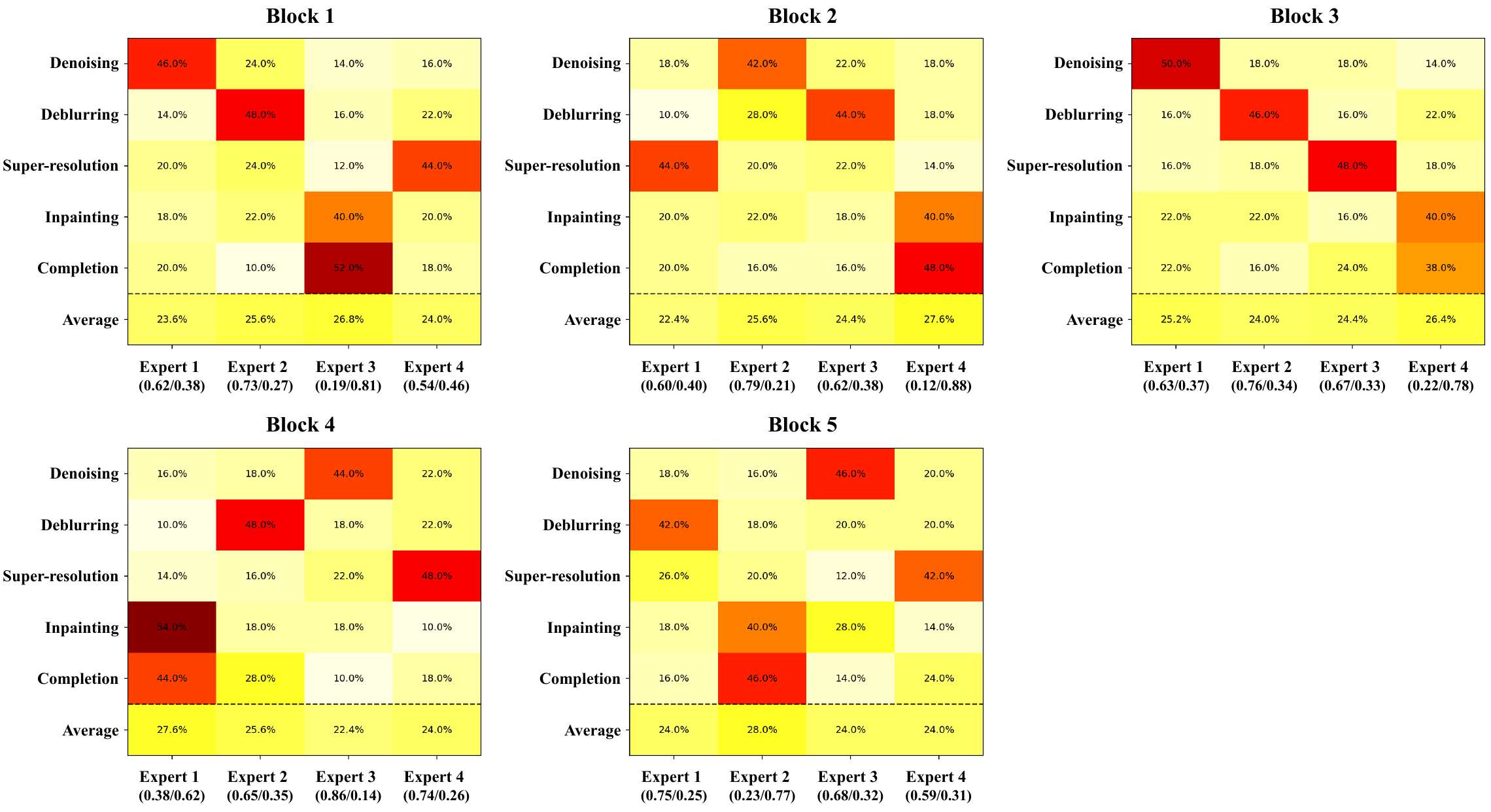}
    \caption{Heatmaps of expert activation rates for each MoE block in DAMP. The five subplots represent 5 distinct blocks, where the vertical axis shows five degradation types and average activation, and the horizontal axis denotes experts with different ($\lambda_s$ / $\lambda_c$). Results demonstrate that data with different degradation types have distinct expert preferences, with all experts maintaining roughly balanced overall activation rates.
}
    \label{fig:expert_activation_heatmap}
\end{figure*}

To validate our claim of expert specialization in the DAMoE module, we extract the learned spatial and spectral weighting parameters ($\lambda_s$ / $\lambda_c$) and plot expert activation heatmaps across all 5 MoE blocks after training. The visualization (Fig.~\ref{fig:expert_activation_heatmap}) confirms our core hypotheses regarding expert disentanglement and degradation-aware routing.

Without any manual assignment or explicit supervision, the learnable weighting parameters naturally polarize into two distinct groups: spatial-heavy experts and spectral-heavy experts. For example, Expert 2 in Block 1 acts as a spatial-heavy expert with $\lambda_s = 0.73$, while Expert 3 in the same block is a spectral-heavy expert with $\lambda_c = 0.81$. This clear polarization is consistently observed across all MoE blocks: Expert 4 in Block 2 achieves $\lambda_c = 0.88$, and Expert 1 in Block 5 achieves $\lambda_s = 0.75$.

Furthermore, the activation heatmaps reveal a strong correlation between degradation type and expert selection. Tasks that primarily corrupt spatial details (e.g., pure deblurring) predominantly activate spatial-heavy experts. Specifically, Expert 2 handles 48\% of deblurring samples in Block 1 and 46\% in Block 3. In contrast, degradations that rely heavily on spectral correlation (e.g., band completion, image inpainting) strongly trigger spectral-heavy experts, with Expert 4 in Block 2 dominating these tasks.

Notably, the "Average" row in the heatmaps confirms that despite task-specific specialization, the overall activation load remains balanced across all four experts (approximately 25\% each). This conclusively demonstrates that our DAMoE routing mechanism effectively disentangles complex hyperspectral image restoration into targeted, non-redundant spatial and spectral sub-tasks.

\subsection{Computational Scalability with Spectral Resolution}
\label{sec:appendix_computational_scalability}

We analyze the computational efficiency of our DAMP framework with respect to spectral resolution. Both our Transformer backbone and Degradation Prompt (DP) extraction module are mathematically designed to be efficient in the spectral dimension. To empirically validate this, we test the inference overhead of DAMP across spectral band counts ranging from 16 to 512, with the spatial resolution fixed at $512\times 512$.

As shown in Table~\ref{tab:computational_overhead_spectral_bands}, the computational cost scales gracefully rather than exponentially with increasing spectral resolution. For example, increasing the number of bands from 31 to 256 only increases the inference time by approximately 1.47$\times$ (from 133.18 ms to 195.81 ms). Critically, the number of model parameters increases marginally (only 0.26 M from 31 to 256 bands) because the core convolution and attention weights are band-agnostic. This demonstrates the excellent efficiency and scalability of DAMP for processing dense hyperspectral data.

\begin{table}[htbp]
    \centering
    \caption{Computational overhead across different spectral band counts.}
    \label{tab:computational_overhead_spectral_bands}
    \begin{adjustbox}{width=0.7\linewidth}
    \begin{tabular}{c c c c}
    \toprule
    Band Counts & FLOPs (G)↓ & Params (M)↓ & Time (ms)↓ \\
    \midrule
    16 & 309.20 & 14.90 & 129.69 \\
    31 & 313.80 & 14.91 & 133.18 \\
    100 & 334.96 & 14.99 & 148.54 \\
    256 & 382.81 & 15.17 & 195.81 \\
    512 & 461.33 & 15.47 & 293.07 \\
    \bottomrule
    \end{tabular}
    \end{adjustbox}
\end{table}

\subsection{Evaluation in Realistic Scenarios}
\label{sec:appendix_realistic_scenarios}

To further validate the generalization ability of our DAMP framework in practical applications, we conduct additional experiments on real-world hyperspectral data. Following the experimental protocol of LDERT~\cite{li2024hsidiff}, we evaluate our method on the realistic dataset introduced in~\cite{zhang2022guided}. We randomly select 8 images for testing and use the remaining images for model training.

The quantitative results are summarized in Table~\ref{tab:realistic_scenarios}. Our proposed DAMP method demonstrates superior performance, outperforming all compared baseline methods across all three evaluation metrics in the real-world setting.

\begin{table}[htbp]
    \centering
    \caption{Performance comparison on real-world hyperspectral dataset.}
    \label{tab:realistic_scenarios}
    \begin{adjustbox}{width=0.7\linewidth}
    \begin{tabular}{l c c c}
    \toprule
    Method & PSNR (dB)↑ & SSIM↑ & SAM↓ \\
    \midrule
    PromptIR & 18.45 & 0.442 & 14.315 \\
    MP-HSIR & 18.41 & 0.486 & 19.052 \\
    \textbf{DAMP (Ours)} & \textbf{20.33} & \textbf{0.542} & \textbf{12.443} \\
    \bottomrule
    \end{tabular}
    \end{adjustbox}
\end{table}

\end{document}